\documentclass[lettersize,journal,twoside]{IEEEtran}
\usepackage{amsmath,amsfonts}
\usepackage{algorithmicx}
\usepackage{algorithm}
\usepackage{algpseudocode}
\usepackage{array}
\usepackage{balance}
\usepackage{booktabs}
\usepackage{caption}
\usepackage{color}
\usepackage{diagbox}
\usepackage{enumerate}
\usepackage{graphicx}
\usepackage[hidelinks]{hyperref}
    \hypersetup{
        colorlinks=true,            % 激活链接颜色，去掉链接边框
        linkcolor=red,              % 文档内部链接颜色（如图表等引用）
        citecolor=blue,             % 文献引用链接颜色
        filecolor=mycustompurple,   % 文件链接颜色
        urlcolor=magenta            % 外部URL链接颜色
    }
\usepackage[utf8]{inputenc} % allow utf-8 input
\usepackage{mathrsfs}
\usepackage{microtype}      % microtypography
\usepackage{multirow}
\usepackage[numbers]{natbib}
\usepackage{nicefrac}
\usepackage{stfloats}
\usepackage[caption=false,font=scriptsize,labelfont=rm,textfont=rm]{subfig}
\usepackage{tabularx}
\usepackage{textcomp}
\usepackage{threeparttable}
\usepackage{times}
\usepackage{titlesec}
    \titleformat{\subsubsection}[runin]{\normalfont\normalsize\bfseries}{}{.5em}{}
    \titleformat{\paragraph}[runin]{\normalfont\normalsize\bfseries}{}{.02em}{}
\usepackage{url}
\usepackage{verbatim}
\usepackage[dvipsnames]{xcolor}
    \definecolor{mycustompurple}{RGB}{154, 36, 79} % 定义自己的颜色

\begin{document}

%\title{Towards Accurate Palm-Vein Recognition: Hybridizing Mamba and CNN with Neural Architecture Search}

\title{Neural Architecture Search based Global-local Vision Mamba for Palm-Vein Recognition}

\author{Huafeng Qin*, Yuming Fu*, Jing Chen, Mounim~A.~El-Yacoubi,  Xinbo Gao, ~\IEEEmembership{Fellow, IEEE}, and Feng Xi     
        % <-this % stops a space
\thanks{H. Qin, Y. Fu, J. Chen, F.Xi, and Y. Xin are with the School of Computer Science and Information Engineering, Chongqing Technology and Business University, Chongqing 400067, China (e-mail: qinhuafengfeng@163.com, fym291715@163.com, jcccchen@163.com).} %\thanks{Y. Li is with the College of Computer Science, Chongqing University, Chongqing 400044, China (e-mail: yantaoli@cqu.edu.cn). }
\thanks{M. A. El-Yacoubi is with SAMOVAR, Telecom SudParis, Institut Polytechnique de Paris, 91120 Palaiseau, France (e-mail: mounim.el\_yacoubi@telecom-sudparis.eu).}
\thanks{ X. Gao is with the Chongqing Key Laboratory of Image Cognition, Chongqing University of Posts and Telecommunications, Chongqing 400065, China (e-mail: gaoxb@cqupt.edu.cn).}
%\thanks{ J. Wang is with the College of Computer Science, China University of Mining and Technology, Jiangsu 221116, China (e-mail:WJ999LX@163.com).}
\thanks{Manuscript received September XX, 2023; revised XXXX XX, 201X. 
%This work was supported in part by %the Scientific Innovation 2030 Major Project for New Generation of AI (Grant No. 2020AAA0107300),  
%the National Natural Science Foundation of China (Grant Nos. 61976030, 62072061 and U20A20176)
%, the Fellowship of China Post-Doctoral Science Foundation (Grant No. 59676651E), the Science Fund for Creative Research Groups of Chongqing Universities (Grant No. CXQT21034), the Chongqing Talent Program (Grant No. CQYC201903246), the Scientific and Technological Research Program of Chongqing Municipal Education Commission (Grant No. KJQN201900848), the Key Project of Science and Technology Research Program of Chongqing Education Commission of China (Grant NO. KJZD-K201901105). 
( Corresponding author: Huafeng Qin.$\star$ Equal contribution.)} 
\thanks{Manuscript received April 19, 2021; revised August 16, 2021.}}

% The paper headers
\markboth{Journal name}%
{Qin \MakeLowercase{{et al.}} : AutoConvMamba: Searching Global and Local Vision Mamba for Palm-Vein Recognition}

% \IEEEpubid{0000-0000/00\$00.00~\copyright~2021 IEEE}
% % Remember, if you use this you must call \IEEEpubidadjcol in the second
% % column for its text to clear the IEEEpubid mark.

\maketitle
\begin{abstract}
 Thanks to its key biometric features, namely high security, high privacy, and liveness recognition, vein recognition has received wide attention recently, with deep learning (DL) models unsurprisingly dominating the field. In particular, Mamba, a recent DL architecture showing robust feature representation with linear computational complexity, has been applied successfully for visual tasks. Vision Mamba, however, captures long-distance feature dependencies but deteriorates local feature details. Besides, manually designing Mamba architecture based on human prior knowledge is very time-consuming and error-prone. To address these limitations, we propose in this paper first a hybrid network structure named Global-local Vision Mamba (GLVM) to learn both local correlations and global dependencies within images for vein feature representation. Second, we design a Multi-head Mamba to learn the dependencies along different directions, so as to improve feature representation of vision Mamba.  Third, to learn complementary features, we propose a ConvMamba block consisting of three branches, Multi-head Mamba branch (MHMamba), Feature Iteration Unit branch (FIU), and Convolutional Neural Network (CNN) branch, with FIU aiming to fuse convolutional local features with Mamba global representations. Finally, we propose a Global-local Alternate Neural Architecture Search (GLNAS) method to search the GLVM optimal architecture  alternately with an evolutionary algorithm. We have carried out rigorous experiments on three public palm-vein datasets to assess performance. The results demonstrate that our approach outperforms representative approaches and achieves state-of-the-art recognition accuracy.
\end{abstract}


\section{Introduction}
Biometrics, one of the most highly-secure security access technology, accurately recognizes individuals by analyzing their physiological or behavioral characteristics, such as face \cite{139758}, fingerprint \cite{587996}, iris \cite{1262028}, eye movement \cite{2024EmMixformer}, \textit{etc.} Compared to traditional biometrics, vein recognition has attracted wider attention thanks to its higher security and privacy \cite{li2023transformer}. Vein biometrics relies on the nature of the deoxygenated hemoglobin in the blood that absorbs near-infrared light for image collection. When near-infrared light with a wavelength of 700nm-1000nm passes through our body, the hemoglobin absorbs near-infrared light. A camera can capture, therefore, vein patterns as the vein vessels in the image are darker than surrounding areas, resulting in unique vein feature information of any individual.
\subsection{Motivation}
Various approaches have been investigated for vein recognition, based on handcrafted descriptors, \textit{i.e.},  maximum Curvature \cite{2007Extraction}, repeated line tracking \cite{liu2013algorithm}, LLBP \cite{ liu2017customized}, and compact multi-representation \cite{ 9772684} to extract vein features. These approaches, however, are not only highly dependent on the designer's experience and prior knowledge, but their representation capacity is limited, resulting in poor performance. 

Recently, deep learning (DL) neural networks (NN) have shown robust feature representation capacity and achieved state-of-the-art performance in various tasks such as image classification \cite{He_2016_CVPR, dosovitskiy2020image} and data augmentation \cite{iclr2024adautomix, qin2024sumix}. Subsequently, various DL approaches have been proposed for vein recognition \cite{2024StarLKNet, das2018convolutional, yang2019fv, yang2020fvras, pan2020multi}, that outperformed traditional approaches, whether handcrafted \cite{2007Extraction, liu2013algorithm, liu2017customized} or traditional machine learning (ML)-based \cite{veluchamy2017system, kamaruddin2019new, sulaiman2019unsupervised, wright2010sparse, yang2021finger}.

DL-based vein recognition approaches are mainly divided into two categories: CNN-based vein-based \cite{das2018convolutional, yang2020fvras,9354642,shen2021finger} and Transformer-based \cite{qin2023label, qin2024attention}. Due to their parameters sharing and local perception, CNNs \cite{He_2016_CVPR, yang2020fvras} are convenient for image-based vision tasks, \textit{e.g.} as they model well the 2D structure by capturing spatial local information. As a result, they brought significant improvement in terms of vein recognition accuracy.  Due to their limited local receptive field \cite{li2023local}, however, CNNs overlook global dependencies, which are as important as local dependencies. 
Transformers \cite{vaswani2017attention}, first proposed for Natural Language Processing (NLP), have been successfully introduced into the Computer Vision through Vision Transformers (ViTs) \cite{dosovitskiy2020image} with impressive performance. Thanks to their attention mechanism, Transformers, capable of capturing long-term dependencies between different positions in an input sequence, have shown excellent performance for vein recognition.  
Nevertheless, Transformers' self-attention mechanism results in low inference speed and huge memory usage when processing long-sequence visual tasks \cite{zhu2024vision}. Recently, State Space Models (SSMs)\cite{gu2022efficiently} have shown great potential for long sequence modeling with linear complexity. These models can be seen as a combination of Recurrent Neural Networks (RNNs) and Convolutional Neural Networks (CNNs).
Some researchers incorporated time-varying parameters into the SSM and proposed a hardware-aware algorithm named Mamba \cite{gu2023Mamba} for more efficient training and inference. Inspired by Mamba's success \cite{gu2023Mamba}, a novel variant has been investigated for image classification \cite{zhu2024vision, huang2024localMamba, liu2024vision}, segmentation \cite{wang2024Mamba,xing2024segMamba} and synthesis \cite{teng2024dim}, with comparable or superior performance \emph{w.r.t} CNNs and ViTs \cite{dosovitskiy2020image}. Mamba's superior scaling performance makes it an alternative foundation model to Transformers \cite{zhu2024vision}.
Unfortunately, there are some limits to Mamba architectures and DL-based models for vein recognition tasks:
\begin{itemize}
\item Although Mamba has been successfully applied for processing natural images and videos, a generic pure-SSM-based backbone network has not been explored yet for biometrics, \textit{e.g.} vein recognition. Besides, as described in \cite{zhu2024vision, gu2022efficiently}, the SSMs exhibit linear or near-linear computational complexity with sequence length and have principled mechanisms for modeling long-term dependencies, making them particularly suited for handling long-sequence data. Similar to ViTs, however, Mamba may ignore local relationships and structure information inside a patch, which are as important as global dependencies \cite{peng2023conformer}. It is worth, therefore, to explore a mechanism to take both compensation features for enhanced representation learning of Mamba. 
\item In most vein recognition works, a DL architecture is manually designed by human experts, thereby suffering from the following drawbacks.  First, this manual design is very time-consuming as there are too many factors to consider. Second, as a huge candidate network architecture space exists, experts are not able to explore it, resulting in finding sub-optimal hyperparameters only. Third, these manually designed architectures are not generalizable to other datasets or more complex tasks, thereby limiting their application in the real world. To address this problem, Neural Architecture Search (NAS) technology \cite{zoph2016neural, real2017large, bender2018understanding, guo2020single}  shows great potential in automatically searching network architectures. Jia \emph{et.al} \cite{jia20212d} introduced NAS for 2D and 3D palmprint recognition and palm vein recognition. Subsequently, an Attention Gated recurrent unit-based Neural Architecture Search (AGNAS) \cite{qin2023ag} was proposed for the same purpose for finger-vein recognition tasks. The mainstream fast Neural Architecture Search methods \cite{guo2020single,qin2023ag,chen2021autoformer} samples candidate sub-networks from the supernet during each training iteration, with the optimal sub-network retained at the end of training. These solutions, however, fail to work well on huge search spaces as there is a low correlation between the retrained subnet and the subnet sampled from the supernet.
\end{itemize}

\subsection{Our work}
 To address these limitations, our paper proposes a Global-local Vision Mamba (GLVM) to capture local and global features for vein recognition. GLVM is stacked by multiple ConvMamba modules, each comprising three crucial branches: CNN branch, Multi-head Mamba (MHMamba) branch, and Feature Interaction Unit branch. The CNN branch is responsible for extracting the local details while the MHMamba branch aims to capture the global dependencies' representation, with the Feature Interaction Unit branch combining the convolutional local features with MHMamba global representation, thereby allowing the proposed GLVM to learn a comprehensive robust vein feature representation.  
 To automatically search our GLVM architecture, we propose a novel Global-local Neural Network Search (GLANAS) strategy to find the optimal network. GLANAS is split into two sub-spaces: global search space and local search space. We alternately search global and local hyperparameters with an evolutionary algorithm to obtain the optimal architecture. Compared to traditional search strategies, our proposed GLANAS stabilizes the search process and improve search performance. To summarize, the main contributions of our paper are as follows:
\begin{itemize}
    \item To our best knowledge, this work is first to accommodate the Mamba model for vein recognition tasks and investigates a Global-local vision Mamba (GLVM) to enhance vein feature learning capacity. 
   \item We propose a Multi-head Mamba (MHMamba) module with a multi-direction scanning mechanism to improve feature representation capacity. Our GLVM is developed by integrating a CNN branch, MHMamba branch and a Feature Interaction Unit branch to learn conjointly local features and global representations for vein recognition.
    \item We are the first to improve Mamba architecture automatically by NAS for vein recognition. We design a Global-local Alternating Neural Network Search algorithm (GLANAS) to find the optimal architecture hyperparameters of the proposed GLVM.   
    \item We have conducted rigorous experiments on three public palm-vein datasets to assess the performance of the proposed methods. The results show that our methods achieve the highest performance w.r.t the state of the art. 
\end{itemize}
\section{Related Works}
\subsection{Traditional Vein Recognition Algorithms}  \label{trad}
Traditional vein recognition algorithms can be divided into two categories: 
(1) Handcrafted algorithms, relying on manually designed descriptors to extract vein texture features, such as curvature-based methods \cite{2007Extraction, yang2017finger}, local binary pattern-based methods \cite{kang2014contactless, liu2017customized}, Gabor filter-based methods \cite{han2012palm, wang2023residual}, \textit{etc.}. As they are highly based on prior assumption, they are unable to represent complex venous features. 
(2) Traditional machine learning algorithms, relying on shallow learning algorithms to model vein features, including SVM \cite{veluchamy2017system}, PCA \cite{kamaruddin2019new}, K-means \cite{sulaiman2019unsupervised}, Sparse Representation (SR) \cite{wright2010sparse}, Low-Rank Representation (LRR) \cite{yang2021finger}. As their expression ability is limited, they are not well adapted to complex venous data.

\subsection{Deep learning-based Vein Recognition Algorithms}   \label{dlvr}
Thanks to their strong representation ability, deep multi-layer neural networks automatically learn features from a large amount of complex high-dimensional data. Das et al.\cite {das2018convolutional}, for instance, proposed a CNN for finger vein recognition. %fvcnn 2018
Yang et al. \cite{yang2020fvras} then proposed a unified CNN that integrated conjointly vein recognition and anti-spoofing. % FVRASNet
Qin et al. \cite{9354642} proposed a CNN and a multi-scale multi-direction GAN-based data augmentation model for single-sample palm vein recognition, %pvcnn 2021
while Shen et al. \cite{shen2021finger} proposed a lightweight finger vein CNN recognizer to reduce computational power consumption.
More recently, Qin et al. proposed a Label-Enhancement Multi-Scale vision Transformer for palm vein recognition \cite{qin2023label}, and an Interactive Vein Transformer (IVT) with an Attention-based Label Enhancement (ALE) scheme to learn the label distribution in vein classification tasks \cite{qin2024attention}.

\subsection{Mamba-based Algorithms}
In 2023, Gu et al. \cite{gu2023Mamba} proposed Mamba, an end-to-end network architecture with a Selective State Space Model, ensuring fast inference and linear complexity, outperforming Transformer's performance on many sequence tasks. In 2024, Zhu et al. \cite{zhu2024vision} introduced Vision Mamba for computer vision.  Mamba's bidirectional scanning mechanism enables it to achieve prominent performance on image classification and semantic segmentation tasks. Subsequently, various variants of Vision Mamba appeared, such as VMamba introducing a cross-scan module to extract global features \cite{liu2024vmamba} and LocalVim, introducing a local window scanning strategy to capture local features, while maintaining the original global dependency representation ability of Mamba \cite{huang2024localMamba}.

\subsection{Neural Architecture Search Algorithms}    \label{nasa}
Neural Architecture Search (NAS) is an automated network architecture learning approach to reduce the cost of heavy network design and improve model's performance on specific tasks.
Bergstra et al. \cite{bergstra2012random} proposed a random search algorithm to uniformly sample within a given range of hyperparameters, and Zoph et al. \cite{zoph2016neural} subequently proposed a reinforcement learning-based NN architecture search method.
In 2020, Guo et al. \cite{guo2020single} proposed a single-path, one-shot NAS method using uniform sampling. The method simplifies the NAS process by training a supernet that can efficiently sample and evaluate subnets, solving the slow convergence and high computational cost associated with traditional NAS techniques.
In 2021, Chen et al. \cite{chen2021glit} proposed a hierarchical network search algorithm with an evolutionary algorithm to search for the optimal Vit architecture from two levels, while Chen et al. \cite{chen2021autoformer} proposed a weight entanglement strategy that allows different candidate Transformer blocks to share weights, which enables more efficient searching of Vit architectures.
In 2022, Su et al. \cite{su2022vitas} developed a new cyclic weight-sharing mechanism for token embedding in Vit, while Qin et al. \cite{qin2023ag} proposed, in 2023, a NAS method based on attention gated recurrent unit to automatically search the optimal network architecture for finger vein recognition tasks for the first time.

\section{Mamba Preliminaries}

\subsection{State Space Models}
The State Space Model (SSM) \cite{gu2021combining} is a mathematical model describing and analyzing the behavior of dynamical systems. In deep learning, SSMs are used for sequential data processing, such as time series analysis, natural language processing and video understanding. Long-term dependencies in the data can be better captured by mapping sequential data into a state space. Given a one-dimension input sequence $x(t)\in\mathbb{R}^{D}$ and output $y(t)\in\mathbb{R}^{D}$, SSMs use a set of equations to represent the relationship between them, as shown in Eq. (\ref{1}).
\begin{equation}
\label{1}
\begin{aligned}
h'(t)=&\, \textbf{A}h(t)+\textbf{B}x(t), \\
y(t)=&\, \textbf{C}h(t), 
\end{aligned}
\end{equation}
where $h(t)\in\mathbb{R}^{V}$ is a state space storing the system's state at time $t$ and $V$ is its size, $\mathbf{A}\in\mathbb{R}^{D \times V}$ is a state transition matrix, describing how the state changes with time. $\mathbf{B}\in\mathbb{R}^{D \times 1}$ is the input control matrix, describing how the external input affects the change of the state space. $\mathbf{C}\in\mathbb{R}^{D \times 1}$ is the output matrix, describing how the state space is mapped to the output space.

As it is continuous, this system is first discretized through a zero-order hold (ZOH) as shown in Eq. (\ref{2}), with $\boldsymbol{\Delta}\in\mathbb{R}^{D} > 0$ a parameter representing the step size to transform the continuous parameters ($\mathbf{A}$,$\mathbf{B}$) into a discrete form ($\overline{\mathbf{A}}$, $\overline{\mathbf{B}}$).
\begin{equation}\begin{aligned}
\label{2}
&\overline{\mathbf{A}}=e^{\Delta \mathbf{A}},\\
&\overline{\mathbf{B}}=(\Delta \mathbf{A})^{-1}(e^{\Delta \mathbf{A}}-I)\cdot\Delta \mathbf{B}.
\end{aligned}\end{equation}
After the discretization, Eq. (\ref{1}) is rewritten as Eq. (\ref{3}):
\begin{equation}
\label{3}
\begin{aligned}
h'(t)=&\, \overline{\mathbf{A}}h(t)+\overline{\mathbf{B}}x(t), \\
y(t)=&\, \mathbf{C}h(t).
\end{aligned}
\end{equation}
The sequence operation of the state space model is then transformed into a convolution operation by designing an appropriate convolution kernel, as shown in Eq. (\ref{4}):
\begin{equation}\begin{aligned}
\label{4}
\overline{\mathbf{K}}&=\left(\mathbf{C}\bar{\mathbf{B}},\mathbf{C}\bar{\mathbf{A}}\bar{\mathbf{B}},\cdots,\mathbf{C}\bar{\mathbf{A}}^q\bar{\mathbf{B}},\cdots, \mathbf{C}\bar{\mathbf{A}}^{Q-1}\bar{\mathbf{B}}\right),\\
y&=x*\overline{\mathbf{K}},
\end{aligned}\end{equation}
where $\overline{\mathbf{K}}\in\mathbb{R}^{Q}$ represents the convolution kernel of SSM. This greatly speeds up the computation of the SSM as the convolution can be computed in parallel.

\subsection{Selective State Space Models} \label{S6}
The parameters ($\mathbf{A, B, C}$) of the traditional SSM (usually refers to \textbf{S4} \cite{gu2022efficiently}) are linear time-invariant, \textit{i.e.}. Due to this limitation, $\textbf{S4}$ cannot perform contextual content awareness and reasoning well. To solve this problem, Selective State Space Model (named \textbf{S6} or \textbf{Mamba} \cite{gu2023Mamba}) appeared, introducing a dynamic selection mechanism that makes the $\mathbf{B}$, $\mathbf{C}$, and $\boldsymbol{\Delta}$ depend on the input, so that the model can adaptively adjust its parameters according to the input. Given the input sequence $x\in\mathbb{R}^{L \times D}$, $\mathbf{A}\in\mathbb{R}^{D \times V}$ consistent with S4, Mamba calculates $\mathbf{B}\in\mathbb{R}^{L \times V}, \mathbf{C}\in\mathbb{R}^{L \times V}, \boldsymbol{\Delta}\in\mathbb{R}^{L \times D}$ by Eq. (\ref{5}):
\begin{equation}\begin{aligned}
\label{5}
\mathbf{B} &= \text{Linear}_B(x) \\ 
\mathbf{C} &= \text{Linear}_C(x) \\
\boldsymbol{\Delta} &= \text{SoftPlus}_\Delta(\text{Linear}_\Delta(x) + \text{bias})
\end{aligned}\end{equation}
\begin{figure*}[ht!]
    \centering
    \includegraphics[width=0.8\linewidth]{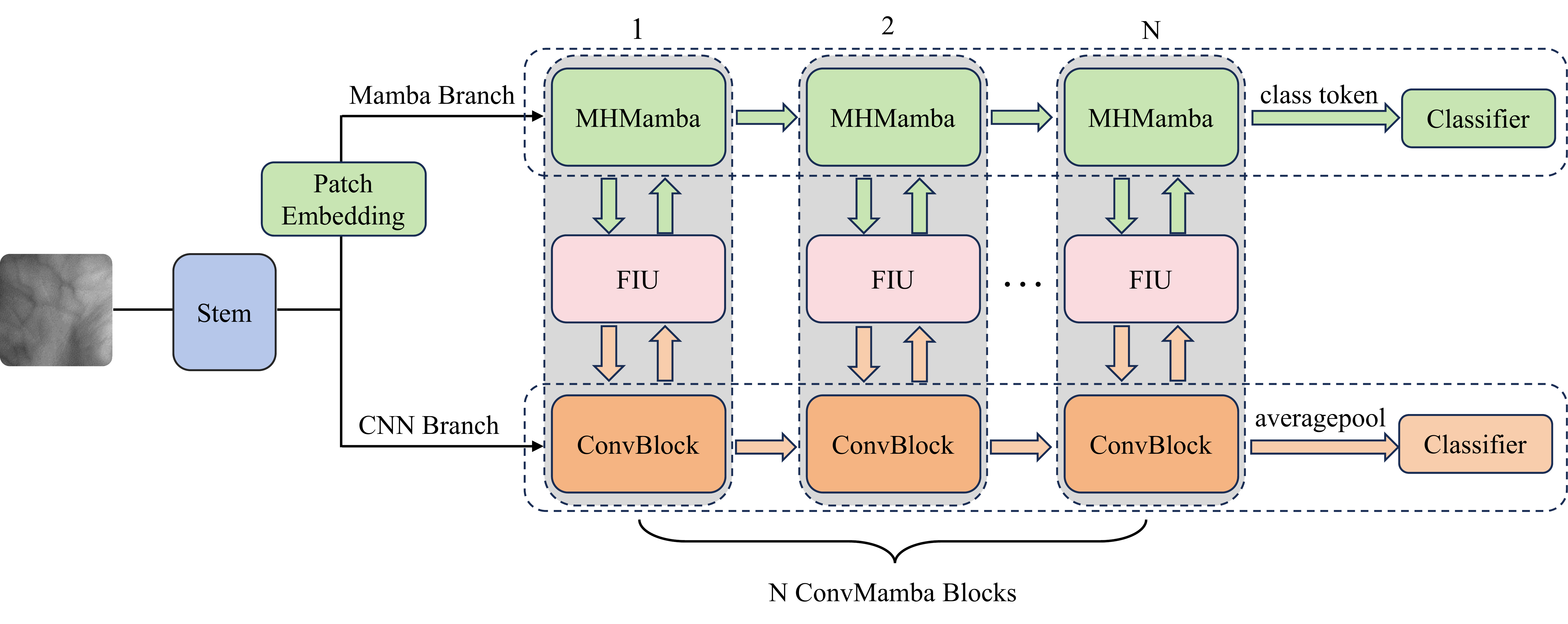}
    \caption{The framework of the proposed Global-local Vision Mamba (GLVM). The Global-local Vision Mamba employs a dual-branch hybrid architecture which consists of a stem block, a patch embedding layer, $N$ ConvMamba blocks, and two classifiers.}
    \label{framework}
\end{figure*}

\section{Methodology}
\subsection{Global-local Vision Mamba Framework} \label{convmamba}
Convolutional Neural Networks (CNNs) are good at extracting local features, but fail to capture global representations due to their local receptive fields  \cite{li2023local,guo2022cmt}.  
Thanks to its state space model structure, Vision Mamba \cite{zhu2024vision,huang2024localMamba} have significantly
larger receptive fields, capturing thereby global representations with long-distance feature dependencies. They have shown,a s a result, better performance than Transformers in some applications. To extract global and local vein feature representation, we propose to combine CNN with the Mamba model by designing a dual-branch hybrid architecture named Global-local Vision Mamba (GLVM). As shown in Fig. \ref{framework}, GLVM includes three branches, \textit{i.e.} CNN, Feature Interaction Unit (FIU), and Multi-head Mamba (MHMamba). The CNN branch includes $N$ ConvBlocks and a classifier while the MHMamba branch includes $N$ MHMamba blocks and a classifier. The Feature Interaction Unit (FIU) branch is employed to fuse the features of the CNN branch and the MHMamba branch. The CNN branch comprises a feature pyramid structure to extract compact features, where the size of the feature map is reduced by half w.r.t the size of the previous stage, and the number of feature channels is gradually increased by a factor of $2$ at each stage in the forward process. The feature maps in the last layer are subjected to the max pooling operationThe MHMamba branch comprises $N$ repeated MHMamba blocks, with the size of feature maps in each layer being the same. The class token in the last layer is forwarded to another classifier layer. In the training phase, we average the CrossEntropy losses from the two classifiers to update the gradient. In the test phase, we sum the two classifiers' prediction scores to obtain the classification output. Our model architecture is detailed as follows.

\subsubsection{(1) Stem.} 
~~In the stem block, a $7 \times 7$ convolution with a stride of 2 is used to preliminarily extract the features. To reduce computational complexity and prevent overfitting, the spatial size of feature maps is reduced by a $3 \times 3$ max pooling layer with a stride of 2. Given an input image $X\in \mathbb{R}^{C_0\times H\times W}$, where $C_0$ is the number of channels, $H$ and $W $ is the width and height, the stem block transforms it from the image space into the feature space  $F_{c}^1$ by Eq. (\ref{stem0}):
\begin{align}
\label{stem0}
F_{c}^1 &= \text{MaxPool}_{3\times 3}(\text{ReLU}(\text{BN}(\text{Conv}_{7\times 7}(X)))),
\end{align}
where $F_{c}^1\in \mathbb{R}^{C\times\frac{H}{4}\times\frac{W}{4}}$, $C$ is the number of output channels, ReLU is the active function, BN is the batch normalization and $\text{Conv}_{k\times k}$ is a convolution operator with a kernel of  $k \times k$.

\subsubsection{(2) Patch Embedding.} \label{patchembeding}
~~The patch embedding layer maps the 2D feature map $F_{c}^1$ into a series of 1D patch embeddings. Feature maps $F_{c}^1$ are divided into $L$ non-overlapping patches with size $ p \times p $ and the number of patches is set to $L=\frac{H}{4p}\times \frac{W}{4p}$. We then compute the embeddings of the resulting patches, by implementing a convolution on different feature maps with $D$ kernels of $p \times p$ at a stride of $p$ by Eq. (\ref{pe}):
\begin{equation}
    \label{pe}
    F_{m}^1 = \text{Transpose}(\text{Flatten}(\text{Conv}_{p\times p}(F_{c}^1))),
\end{equation}
where $F_{m}^1\in \mathbb{R}^{L\times D}$, Transpose denotes the matrix transpose operation, and Flatten operation transforms the 2D matrix into 1D vectors, and $D$ also represents the embedding dimension. Finally, $F_{m}^1$ and $F_{c}^1$ are taken as the input of the first ConvMamba block.

\subsubsection{(3) ConvMamba Block.} \label{c-b}
\begin{figure*}[h!]
    \centering
    \includegraphics[width=0.75\linewidth]{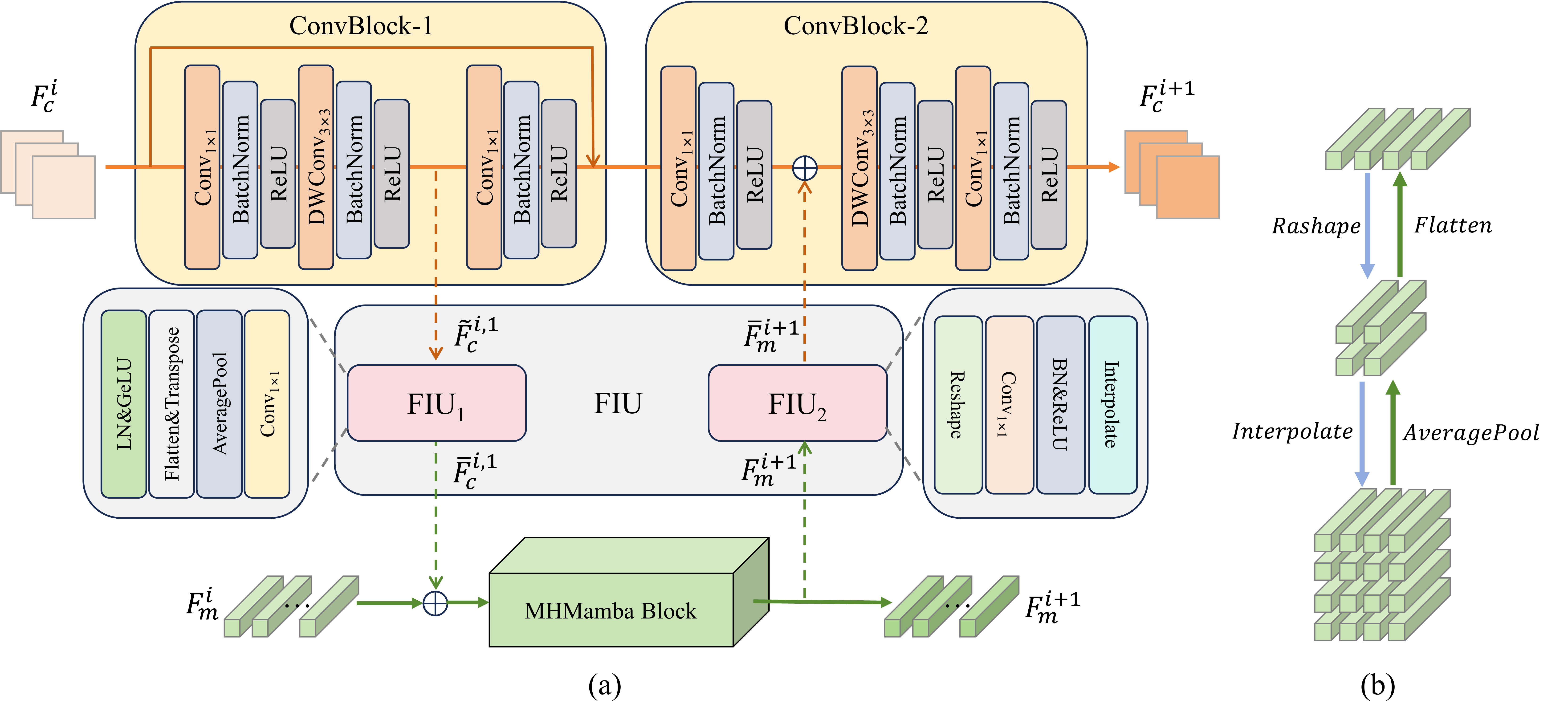}
    \caption{The architecture of ConvMamba block. (a) Each ConvMamba block consists of a MHMamba block, a Feature Interaction Unit, and two sub-convolution blocks, each of which includes $3$ convolutional layers followed by the BatchNorm regularization and ReLU activation. (b) The basic operations in a Feature Interaction Unit, \textit{e.g.}, Flatten and Averagepool.}
    \label{ConvMambaBlock}
\end{figure*}

~~As shown in Fig. \ref{ConvMambaBlock}, the ConvMamba block consists of three branches: CNN, Feature Interaction Unit (FIU), and Multi-head  Mamba (MHMamba), detailed as follows.

\paragraph{CNN Branch.} \label{CNN Branch} 
~~The CNN branch comprises $2$ sub-convolution blocks, each consisting of three convolution layers, followed by batch normalization and ReLU activation. Let $F_{c}^{i}$ be the input feature maps of the CNN branch in the \textit{i}-th ConvMamba block. The forward process is formulated as:\par
 \vspace{-5mm}
\begin{normalsize}
    \begin{alignat}{2}
    & \tilde{F}_{c}^{i,1} &&= \text{DWConv}_{3\times 3}(\text{Conv}_{1\times 1}(F_{c}^{i}))   ,                              \label{c1} \\
    & \hat{F}_{c}^{i,2}   &&= \text{Conv}_{1\times 1}(\text{Conv}_{1\times 1}(\tilde{F}_{c}^{i,1})+F_{c}^{i})   ,            \label{c2} \\
    & \overline{F}_{m}^{i+1} &&= \text{FIU}_2 ({F}_{m}^{i+1})                                ,                           \label{c3} \\
    & F_{c}^{i+1}         &&= \text{Conv}_{1\times 1}(\text{DWConv}_{3\times 3}(\hat{F}_{c}^{i,2} + \overline{F}_{m}^{i+1}))  , \label{c4}
    \end{alignat}
\end{normalsize}%
where DWConv represents depthwise convolution, which effectively reduces computation cost and the number of parameters, and improves model inference speed and efficiency. $\text{FIU}_2$ represents the second module of the FIU, and ${F}_{m}^{i+1}$ is the output of the MHMamba branch in the \textit{i}-th ConvMamba block. For simplification, we neglect the batch normalization and active function in Eqs. (\ref{c1})-(\ref{c4}).\par

\begin{figure*}[ht!]
    \centering
    \includegraphics[width=0.98\linewidth]{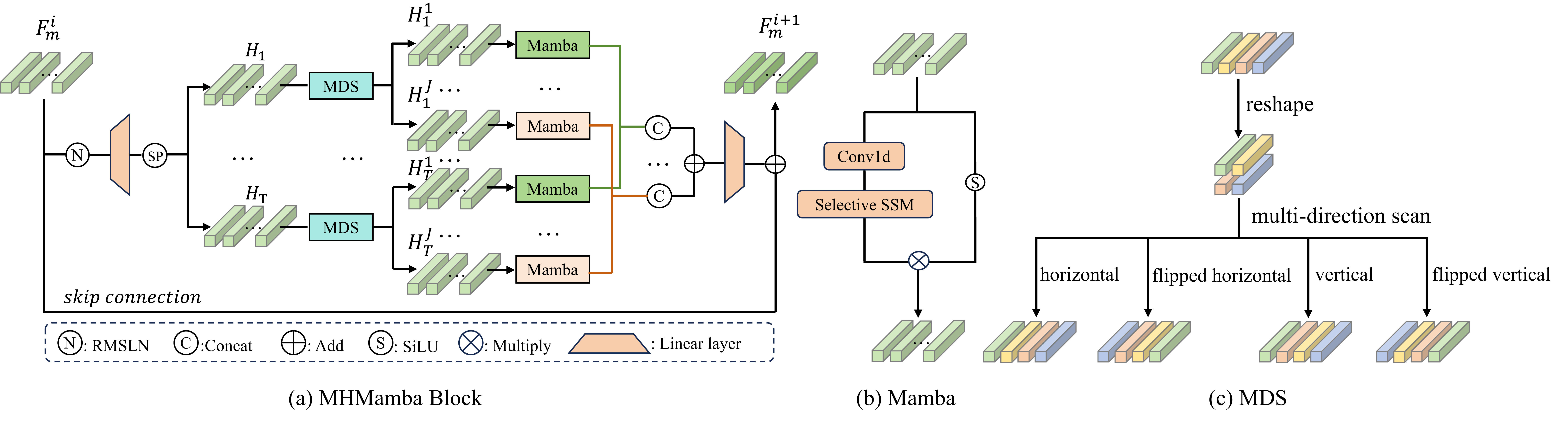}
    \caption{{MHMamba block.(a) The detailed architecture of our MHMamba block, (b) Mamba module in (a), and (c) Multi-direction Scanning (MDS) mechanism in (a), which includes multiple scanning directions, \textit{e.g.}, vertical, flipped vertical, horizontal, and flipped vertical.}}
    \label{MHMambaBlock}
\end{figure*}

\paragraph{Multi-head Mamba Branch.} \label{MHMamba Branch} 
~~To improve the feature representation capacity of the classic Mamba block \cite{gu2023Mamba}, we design a Multi-head Mamba (MHMamba) block as shown in Fig. \ref{MHMambaBlock}. By employing Transformer's \cite{vaswani2017attention} multi-head strategy in addition to a multi-directional scanning strategy, MHMamba greatly improves the recognition accuracy of the Mamba branch. As shown in  Fig. \ref{ConvMambaBlock}(a), ${F}_{m}^{i}$ is the input of MHMamba branch in the \textit{i}-th ConvMamba block and the feature map $\tilde{F}_{c}^{i,1}$ is the intermediate output of the first convolution block of the CNN branch in the \textit{i}-th ConvMamba block. ${F}_{m}^{i}$ and $\tilde{F}_{c}^{i,1}$ are combined by Eq. (\ref{mhmdin}):
\begin{align}
    \tilde{F}_{m}^{i} &= F_{m}^{i} +\text{FIU}_1(\tilde{F}_{c}^{i,1}),  \label{mhmdin}
\end{align}
where $\text{FIU}_1$ is the first module of the feature interaction unit branch. The resulting embedding $\tilde{F}_{m}^{i}\in \mathbb{R}^{L\times D}$ is taken as input of the MHMamba block for the feature extraction and computation (Eqs. (\ref{RMSLN})-(\ref{linear1})). First, we extend the dimension of the input patch embeddings through a linear layer after a prior normalization as shown in Eq. (\ref{RMSLN}):
%\vspace{-4mm}
\begin{align}
     \overline{F}_{m}^{i} = \text{Linear}(\text{RMSLN}(\tilde{F}_{m}^{i})), \label{RMSLN}
\end{align}
{where $\overline{F}_{m}^{i}\in \mathbb{R}^{L\times (E\times D)}$, $E$ is the dimension expansion ratio, RMSLN is the Root Mean Square Layer Normalization \cite{zhang2019root}, and Linear represents the linear projection layer.} Similarly to Transformer, the resulting embeddings $\overline{F}_{m}^{i}$ are then split into different heads $H_{t}\in \mathbb{R}^{L\times D_h}$, $t=1,2,...,T $ (Eq. (\ref{split})):
\begin{align}
    {\Omega}_{m}^{i} = \text{Split}(\overline{F}_{m}^{i}) = \{H_1, H_2, \cdots, H_T\} , \label{split}
\end{align}
where $T$ is the number of heads and $D_h= \frac{E\times D}{T} $ is the dimension of each head. We map all heads $H_{t}$, $t=1,2,...,T$, into $J$ directions (Eq. (\ref{MultiScan})):
\begin{align}
    H_t = \text{MDS}(H_t) = \{H^1_t, H^2_t, \cdots, H^J_t\},  \label{MultiScan}
\end{align}
where MDS is a 2D multi-direction scanning method, which scans the input along different directions,\textit{ e.g.} vertical, flipped vertical, horizontal, and flipped vertical, as shown in Fig. \ref{MHMambaBlock}(c). $H^j_{t}\in \mathbb{R}^{L\times D_h}$ represents the \textit{j}-th direction of the \textit{t}-th head and is further input to the Mamba module (Fig. \ref{MHMambaBlock}(b)) to learn a global feature feature representation (Eq. (\ref{SSM})):
\begin{normalsize}
    \begin{align}
        H^j_t  = \text{SSM}(\text{Conv}_{1 \times1}({H}^{j}_{t})) \odot \text{SiLU}({H}^{j}_{t})  \label{SSM},
    \end{align}
\end{normalsize}%
where SSM is the Selective State Space Model \cite{gu2023Mamba}, SiLU is the activation function \cite{hendrycks2016gaussian}, and $\odot$ represents the element-wise product. We concatenate the heads along the same direction by Eq. (\ref{concat}):
\begin{equation}
    \label{concat}
    \begin{aligned}
        \hat{F}_{m}^{i} &= \text{Concat}\{H^1_1, H^1_2, \cdots, H^1_T\}, \\
                        &+ \text{Concat}\{H^2_1, H^2_2, \cdots, H^2_T\} ,\\
                        &+\cdots+\text{Concat}\{H^J_1, H^J_2, \cdots, H^J_T\} .
    \end{aligned}
\end{equation}
The resulting embeddings $\hat{F}_{m}^{i} $ are forwarded to an another linear layer, computed according to Eq. (\ref{linear1}):
\begin{align}
    {F}_{m}^{i+1} = \text{Linear}(\hat{F}_{m}^{i}).  \label{linear1}
\end{align}

\paragraph{Feature Interaction Unit Branch.} \label{f-a-b}
~~We design a Feature Interaction Unit (FIU) to align the spatial information of the CNN branch's local features and the MHMamba branch's global representations so that they can be fused. As shown in Fig. \ref{ConvMambaBlock}, FIU consists of two sub-modules, \textit{i.e.} $\text{FIU}_1$ and $\text{FIU}_2$.

The $\tilde{F}_{c}^{i,1} \in \mathbb{R}^{C_1\times H_1\times W_1}$ in Eq. (\ref{c1}) is the input of $\text{FIU}_1$ and the  $F_{m}^{i} \in \mathbb{R}^{L\times D}$  in Eq. (\ref{mhmdin}) is the input of MHMamba branch. $\text{FIU}_1$ first aligns the channel numbers of $\tilde{F}_{c}^{i,1}$ and $F_{m}^{i}$  to get $\hat{F}_{c}^{i,1} \in \mathbb{R}^{D\times H_1\times W_1}$ through a $1\times 1$ convolution layer, and then aligns the spatial resolution information of $\hat{F}_{c}^{i,1}$ and $F_{m}^{i}$ to get $\overline{F}_{c}^{i,1} \in \mathbb{R}^{L\times D}$ through an average pool layer followed by flattening and transpose operations. The resulting representation is regularized and activated by LN and GeLU functions \cite{hendrycks2016gaussian}. The forward process of $\text{FIU}_1$ is formulated as:\par
\vspace{-5mm}
\begin{normalsize}
    \begin{align}
        & \hat{F}_{c}^{i,1} = \text{Conv}_{1\times1}(\tilde{F}_{c}^{i,1}), \nonumber \\
        & \overline{F}_{c}^{i,1} = \text{AveragePool}(\text{Transpose}(\text{Flatten}(\hat{F}_{c}^{i,1}) ) )\label{FIU-1}, \\
        & \overline{F}_{c}^{i,1} = \text{GeLU}(\text{LN}(\hat{F}_{c}^{i,1})). \nonumber 
    \end{align}
\end{normalsize}%
The alignment feature representation $\overline{F}_{c}^{i,1}$ will be fused with ${F}_{m}^{i}$ by Eq. (\ref{mhmdin}), as shown in Fig. \ref{ConvMambaBlock}.

{In addition,  the input of $\text{FIU}_2$ is $F_{m}^{i+1} \in \mathbb{R}^{L\times D}$ in Eq. (\ref{linear1}) and the CNN branch's intermediate feature maps is $\hat{F}_{c}^{i,2} \in \mathbb{R}^{C_2\times H_2\times W_2}$ in Eq. (\ref{c2}). 
Similarly, $\text{FIU}_2$ first aligns the channel numbers of $F_{m}^{i+1}$ and $\hat{F}_{c}^{i,2}$ to get $\hat{F}_{m}^{i+1} \in \mathbb{R}^{C_2\times h\times w}$ ($h=w=\frac{H}{4p}=\frac{W}{4p}$) through Transpose, Reshape operations and a $1\times 1$ convolution layer followed by BN regularization and ReLU activation, and then aligns the spatial resolution information of $\hat{F}_{m}^{i+1}$ and $\hat{F}_{c}^{i,2}$ to get $\overline{F}_{m}^{i+1} \in \mathbb{R}^{C_2\times H_2\times W_2}$ by the Interpolate operation. The forward process of $\text{FIU}_2$ is formulated as:}\par
\vspace{-5mm}
\begin{normalsize}
    \begin{equation}
        \label{FIU-2}
        \begin{aligned}
            & \hat{F}_{m}^{i+1} =\text{Conv}_{1\times1}(\text{Reshape}(\text{Transpose}(F_{m}^{i+1}))), \\
            & \hat{F}_{m}^{i+1} =\text{ReLU}(\text{BN}(\hat{F}_{m}^{i+1})), \\
            & \overline{F}_{m}^{i+1} = \text{Interpolate}( \hat{F}_{m}^{i+1}).
        \end{aligned}
    \end{equation}
\end{normalsize}%
The aligned feature representation $\overline{F}_{m}^{i+1}$ is then fused with $\hat{F}_{c}^{i,2}$ (Eq. (\ref{c4})).

\begin{figure*}[h!]
    \centering
    \includegraphics[width=0.7\linewidth]{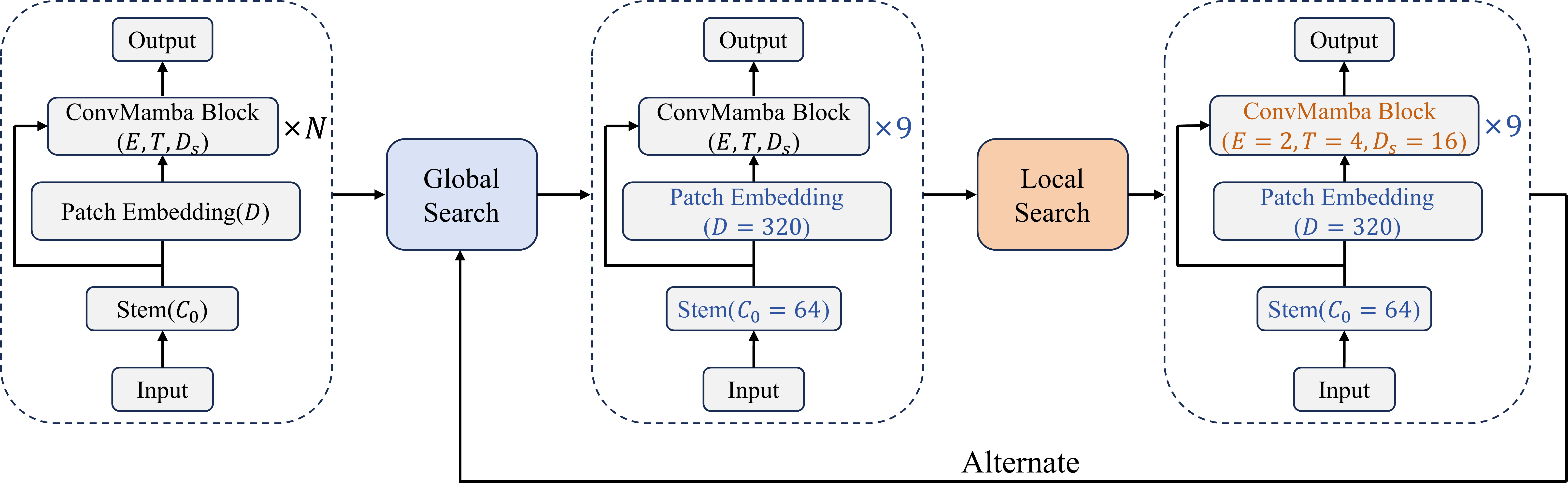}
    \caption{Framework of the GLANAS. We split the search space into two parts, global space and local space, and alternately perform a two-stage search, with each stage of the search employing a weight entanglement strategy to sample subnets from the supernet.}
    \label{anas}
\end{figure*}

\subsection{Search space of Global-local Vision Mamba.}  \label{ssc}
~~The search space consists of the global architecture of Global-local Vision Mamba (GLVM) and the local detail architecture of each Multi-head Mamba (MHMamba) block (Tab. \ref{ss}).

\begin{table}[h!]
\centering
\caption{The search space of our GLVM}
\setlength{\abovecaptionskip}{1pt} %表标题下方间距
\setlength{\belowcaptionskip}{1pt} %表标题上方间距
\renewcommand\arraystretch{1.3}
\scalebox{0.95}{
\begin{tabular}{c|c|c|c}
\toprule[1pt]
\multirow{1}{*}{Stage}  & Hyperparameter &Symbol & (Min, Max, Interval)             \\
\midrule[0.7pt]
\multirow{4}{*}{Global} & Depth          &$N$    & (6, 12, 3)                   \\
                        & Channel Num    &$C$  & (32, 128, 32)                \\
                        & Embed Dim      &$D$    & (128, 384, 64)               \\
                        
\midrule[0.7pt]
\multirow{3}{*}{Local}  & Head Num       &$T$    & (2, 6, 2)                    \\
                        & Expand Ratio   &$E$    & (1, 2, 0.5)                  \\
                        & State Size     &$V$  & (16, 48, 16)                 \\
\bottomrule[1pt]
\end{tabular}
}
\label{ss}
\end{table}

\subsubsection{(1) Global search space.}~~The global architecture hyperparameter determines the scale of the network, which affects the size of all local modules, and plays a key role in the final model performance. Suppose we have $N$ ConvMamba blocks, with $D$ the embedding dimension in the patch embedding layer, and $C$ the number of channels in the stem layer. As shown in Tab. \ref{ss}, $N$, $C$, and $D$ are sampled into discrete values. Let $S_1$, $S_2$, $S_3$ represent the total candidate numbers of hyperparameters $N$, $C$, and $D$ respectively, we can get the search sets $\mathcal{N}_1 = \{N_1, N_2, \cdots, N_{S_1}\}$, $ \mathcal{N}_2 = \{D_1, D_2, \cdots, D_{S_2}\}$, and $ \mathcal{N}_3 = \{C_1, C_2, \cdots, C_{S_3}\}$ of $N$, $C$, $D$ respectively. The final global candidate architecture search set is $\mathcal{N}_G = \mathcal{N}_1 \times \mathcal{N}_2 \times \mathcal{N}_3$, where $\times$ represents the Cartesian product.

\subsubsection{(2) Local search space.}~~The local architecture hyperparameters are associated with the local detail architecture in the MHMamba block (Fig. \ref{MHMambaBlock}(a)). The search space of the detailed architecture comprises the dimension expansion rate $E$, the number of heads $T$, and the size of the state space $V$ in SSM (section \ref{S6}). Similarly, we discretize the continuous space into different values and the candidate numbers in $E$, $T$ and $V$ are $ {S_4}$, ${S_5}$, and ${S_6}$. The candidate sets of $E$, $T$, $V$ are denoted as $ \mathcal{N}_4 = \{E_1, E_2, \cdots, E_{S_4}\}$, $\mathcal{N}_5 = \{T_1, T_2, \cdots, T_{S_5}\}$ and $\mathcal{N}_6 = \{V_1, V_2, \cdots, V_{S_6}\}$ respectively. 
For a MHMamba block $b_n \in \{b_1, b_2, \cdots, b_n, \cdots, b_N\}$, the inner search sets are $\mathcal{N}_4^n$,  $\mathcal{N}_5^n$ and $\mathcal{N}_6^{n}$. The overall search set of block $b_n$ is $\mathcal{S}_n $=$\mathcal{N}_4^n$ $\times $ $\mathcal{N}_5^n$ $\times$ $\mathcal{N}_6^{n}$ and the final local search space is $\mathcal{N}_L = \mathcal{S}_1 \times \mathcal{S}_2 \times  \cdots \times \mathcal{S}_N$.

\subsubsection{(3) Search space size.}~~For each MHMamba block, there are $S_4S_5S_6$ candidate architectures. In our search space, different MHMamba blocks have different architectures. Considering the global and local architecture hyperparameters, the search space size of the global architecture is $S_1S_2 S_3$, the search space size of the local architecture is $(S_4 S_5 S_6)^{N_{S_1}}$, and the final search space contains $S_2 S_3 \sum_{i=1}^{S_1}(S_4 S_5 S_6)^{N_i}$ candidate networks. As listed in Tab. \ref{ss}, $S_1=3$, $S_2=4$, $S_3=5$, $S_4=3$, $S_5=3$, $S_6=3$, and $N_1 = 6, N_2=9, N_{3}=12$, so there are $(S_2 S_3) [(S_4 S_5 S_6)^6+(S_4 S_5 S_6)^9+(S_4 S_5 S_6)^{12}] \approx 3\times10^{18}$ candidate networks, which is a huge search space. The mainstream fast architecture search methods such as one-shot NAS with weight sharing \cite{guo2020single} show poor performance in such a huge search space as it selects one candidate network for weight updating in each training iteration. Due to the huge search space, however, the weights of most candidate architectures are not updated after completing supernet training, resulting in a low correlation between the retrained subnet and the subnet sampled from the supernetwork, making unreliable the architectures searched by using supernet. To solve the problem, we propose a Global-local Alternating Neural Architecture Search (GLANAS) method to get the optimal global and local network architecture with suitable memory requirements.

\subsection{Global-local Alternating Neural Architecture Search} \label{NAS}
Section \ref{ssc} has defined our architecture search space, consisting of global architecture search space and local module search space.
In this section, we design a novel architecture search method named Global-local Alternative Neural Architecture Search (GLANAS). GLANAS contains three stages: global search, local search, and alternative search (Fig. \ref{anas}).

\subsubsection{(1) Global Search.}
~~At the global search stage, we initialize and fix the local architecture hyperparameters in the ConvMamba block to optimize global architecture hyperparameters. The Weight Entanglement method \cite{chen2021autoformer} is applied to find the optimal global architecture. The global search space $\mathcal{N}_G$ is encoded in a supernet $\mathcal{R}(\mathcal{N}_G, W)$, where $W$ denotes the weights of the supernet. Let $\beta$ be the subnets sampled from the supernet $\mathcal{R}$, with $W$ shared among all the candidate architectures $\beta \in \mathcal{N}_G$.  The One-Shot NAS method is usually a two-stage optimization problem. In the first stage, the weight vector $W$ of the supernet $\mathcal{R}$ is optimized by Eq. (\ref{o1}):
\begin{equation}
\label{o1}
    W_{\mathcal{N}_G}=\arg\min_{W}\mathcal{F}_{train}(\mathcal{R}(\mathcal{N}_G,W)),
\end{equation}
where $\mathcal{F}_{train}$ is  the loss function on the training dataset. The one-shot method then samples subnets $\beta$ from the supernet $\mathcal{R}$. The second phase aims to search the optimal subnet $\beta_*$  based on the accuracy of all subnets $\beta\in \mathcal{N}_G$  on the validation set, which is computed by Eq. (\ref{o2}):
\begin{equation}
\label{o2}
    \beta_*=\arg\max_{\beta\in\mathcal{N}_G}\mathrm{~Acc}_{\nu al}\left(\mathcal{R}(\beta,W_{\mathcal{N}_G}(\beta))\right),
\end{equation}
where $W_{\mathcal{N}_G}(\beta)$ is the weight vector of sampled subnet $\beta$  inherited from the supernet weight vector  $\mathcal{R}$, and $\mathrm{Acc}_{\nu al}$ denotes the accuracy of the subnet $\beta$ on the validation dataset. After the global search, we obtain the optimal global architecture hyperparameters $\beta_* = (N_*, C_*, D_*)$, where $*$ represents the optimal choice.

\subsubsection{(2) Local Search.}
~~After the global search stage, we fix the obtained optimal global architecture hyperparameters and optimize the hyperparameters of the local module architectures, \textit{i.e.} all ConvMamba blocks, thanks to the Weight Entanglement method \cite{chen2021autoformer}. Accordingly, we get the optimal local module architecture $\gamma_* =((T^1_{*}, E^1_{*}, V^1_{*}), (T^2_{*}, E^2_{*}, V^2_{*}), \cdots, (T^{N_*}_{*}, E^{N_*}_{*}, V^{N_*}_{*}))$.

\subsubsection{(3) Alternative search.}
~~We have optimized the global search and then local search in a greedy to obtain the global and the local module architectures. Although the latter may be optimal relative to the former, we cannot guarantee, therefore, that the former is optimal relative to the latter. To overcome this issue, we fix the local module hyperparameters to perform global search again, with the final network architecture obtained by alternating this process for $K$ times. 

The proposed global-local alternating search method has the following advantages \textit{w.r.t} existing NAS methods \cite{guo2020single,qin2023ag,chen2021autoformer}. First, the huge search space is split into two smaller ones. Specifically, we divide the original space size is  $S_2 S_3 \sum_{i=1}^{S_1}(S_4 S_5 S_6)^{N_i}$, into two smaller spaces $S_1 S_2 S_3$ and $(S_4 S_5 S_6)^{N_{S_1}}$, with the final search space size being $S_1 S_2 S_3+(S_4 S_5 S_6)^{N_{S_1}} \approx 1.5 \times 10^{17}$. The new search space, therefore, is reduced to less than  $\frac{1}{20}$ of the original one. As the final search space falls into the effective search space range of NAS methods, this makes it easier for the Weight Entanglement method \cite{chen2021autoformer} to obtain a better model. Second, our alternate search scheme ensures that a highly compatible and performant architecture can be found in the huge search space.

\section{Experiments and Results}
To assess our approach performance, we carry out rigorous experiments on three public vein datasets. We compare our approach with six advanced vein recognition models (PVCNN \cite{9354642}, FVCNN \cite{das2018convolutional}, LWCNN \cite{shen2021finger}, FVRASNet \cite{yang2020fvras}, LE-MSVT \cite{qin2023label}, ALE-IVT \cite{qin2024attention}), and four state of the art models (ResNet50 \cite{He_2016_CVPR}, Vim \cite{zhu2024vision}, Vit \cite{dosovitskiy2020image}, LocalVim \cite{huang2024localMamba}). Existing NAS methods, namely Random Search \cite{bergstra2012random}, SPOS \cite{guo2020single}, AutoFormer \cite{chen2021autoformer} and AGNAS \cite{qin2023ag}, are repeated to demonstrate the effectiveness of our proposed GLANAS method.
\subsection{datasets}
\textbf{(1) TJU\_PV}: The Tongji University Palm Vein dataset \cite{zhang2018palmprint} comprises $12,000$ images ($300$ subjects $\times$ $2$ palms $\times$ $10$ images $\times$ $2$ sessions), with an average time interval of about two months. Treating each palm as a category, we get $600$ categories with $20$ images each.

\textbf{(2) VERA\_PV}: The VERA Palm Vein dataset \cite{tome2015vulnerability} comprises $2200$ images ($110$ subjects $\times$ $2$ palms $\times$ $5$ images $\times$ $2$ sessions). Treating each palm as a category, we get $220$ categories with $10$ images each.

\textbf{(3) HKPU\_PV}: The Hong Kong Polytechnic University's multispectral Palm Vein dataset \cite{zhang2009online} consists of $6,000$ images ($250$ subjects $\times$ $2$ palms $\times$ $6$ images $\times$ $2$ sessions), with an average time interval of nine days. If we treat each palm as a category, we get $500$ categories,with $12$ images each. 

\begin{figure}[ht!]
\centering
\subfloat[]{
		\includegraphics[width=0.50\linewidth]{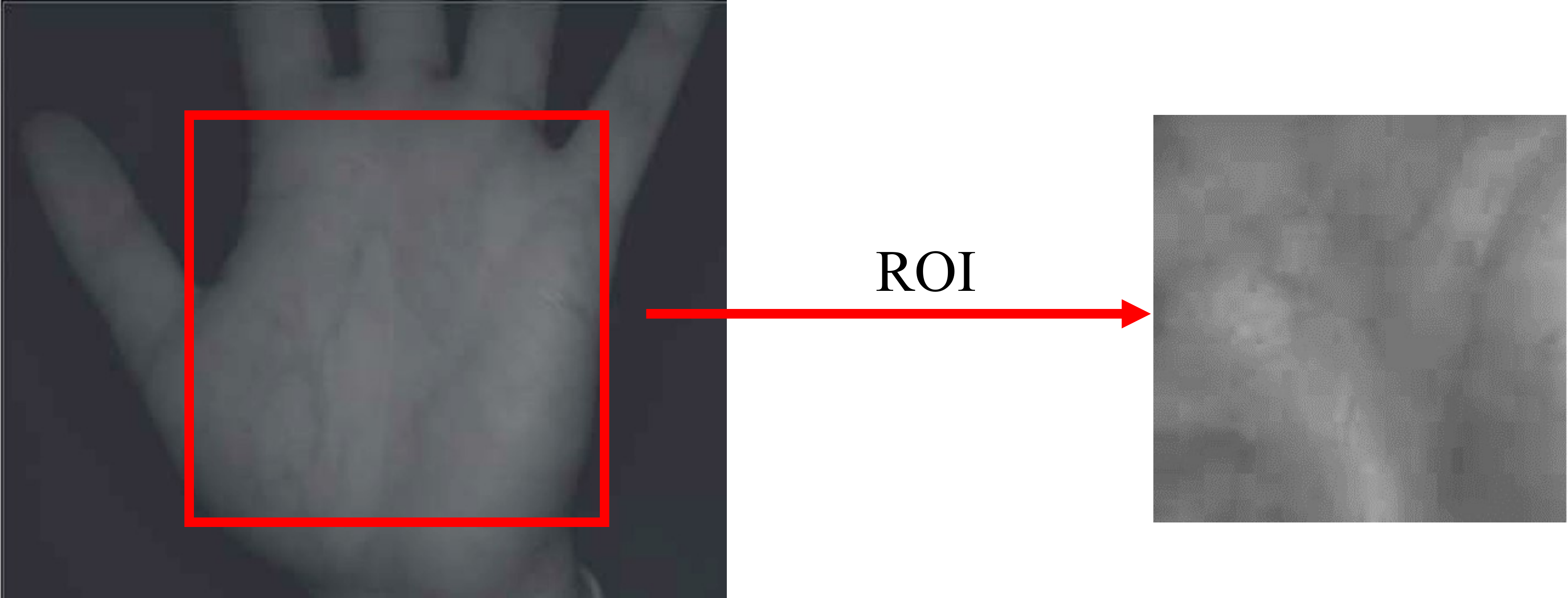}}
\\
\subfloat[]{
		\includegraphics[width=0.50\linewidth]{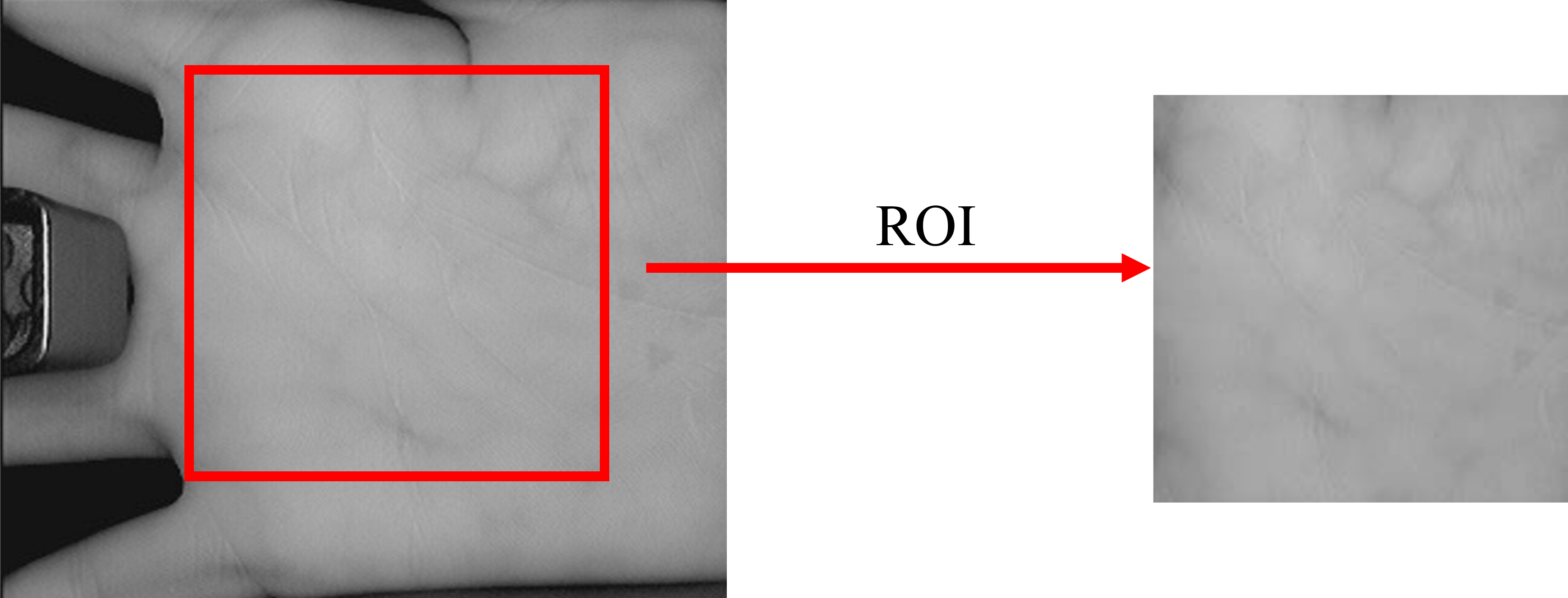}}
\\
\subfloat[]{
		\includegraphics[width=0.50\linewidth]{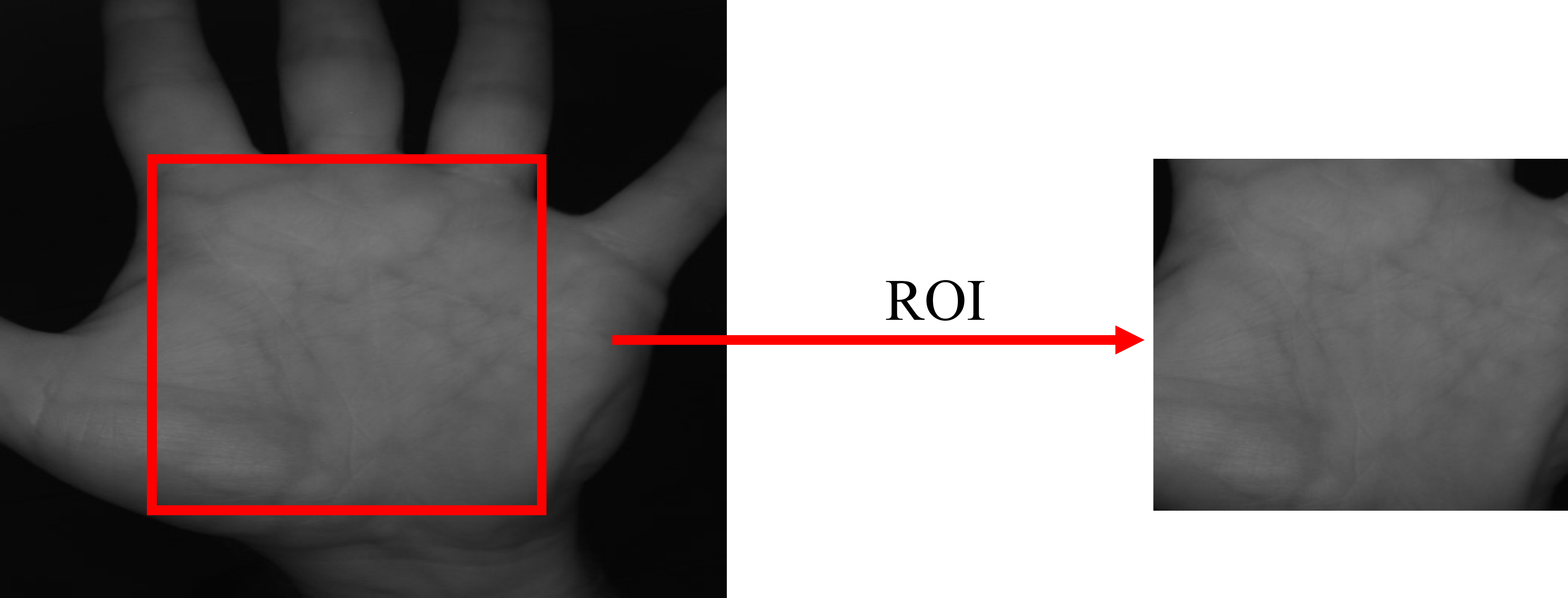}}
\caption{Preprocessing results on three datasets. Original palm image and ROI on (a)  TJU\_PV dataset, (b) HKPU\_PV dataset, and (c) VERA\_PV dataset.}
\label{oriandroi}
\end{figure}

The palm vein images from the three datasets, captured in a contact or contactless way, are usually subject to some translations and rotations, which may cause mismatching. Besides, the background region in captured images do not provide any discrimination information. Each  image's Region of Interest (ROI) is therefore extracted from the original image based on the preprocessing approach \cite{qin2019iterative} and normalized to the same direction, as shown in  Fig. \ref{oriandroi}. 

\subsection{Experimental Settings}
\textbf{(1) dataset Settings.} ~We split each palm dataset into three sets: training set, validation set, and test set, with the validation set used for model tuning. For each dataset, the first session images are used as for training while the second session images are divided into the validation and test sets. Accordingly, TJU\_PV comprises $6000$ images ($600$ classes $\times 10$ images). In the second session, we use for each class the first $5$ for validation and the last $5$ for test, obtaining thereby $3,000$ images ($600$ classes $\times 5$ images) for validation and $3,000$ images ($600$ classes $\times 5$ images) for test. 
Similarly, VERA\_PV comprises $1,100$ i mages ($220$ classes $\times 5$ images) for training, $440$ images ($220$ classes $\times 2$ images) for validation, and $660$ images ($220$ classes $\times 3$ images) for test, while HKPU\_PV comprises $3,000$ images ($500$ classes $\times 6$ images) for training, $1,500$ images ($500$ classes $\times 3$ images) for validation and $1,500$ images ($500$ classes $\times 3$ images) for test. 
Due to vein datasets' limited size, which usually leads to deep learning model overfitting, we perform data augmentation consisting of geometric transformations (rotation, scaling, flipping, cropping), color enhancement transformations (contrast, brightness, hue, and saturation), and the mix-up method \cite{zhang2017mixup}. 

\textbf{(2) Training Hyperparameters.} ~The adaptive moment estimation with weight decay (AdamW) \cite{loshchilov2017fixing} optimizer is used for parameter optimisation. The learning rate is set to $0.001$, with warmup epoch number of $5$, eventually reduced to $0.0001$ with a cosine scheduler. The weight decay is $0.05$, and the minibatch sizes are set to $60$, $50$, and $40$ for resp. TJU\_PV, VERA\_PV, and HKPU\_PV datasets. The maximum number of training epochs is set to $500$. All experiments are implemented on PyTorch with an NVIDIA Tesla A100 80G GPU. 

\textbf{(3) Evaluation Metrics.} ~Top-1 recognition accuracy (ACC) and equal error rate (EER) are used to assess performance. The EER is the error rate when the false accepted rate (FAR) equals the false rejected rate (FRR). The ROC curve provides a comprehensive view of model's performance at all thresholds and is plotted by FAR against the true accepted rate (TAR=1-FRR). The area under the ROC curve quantifies a model's recognition ability, with a larger area implying a better model recognition performance. Thus, we plot ROC curves for all methods to test their performance.

\subsection{Performance of Global-local Vision Mamba}
\begin{figure*}[ht!]
\centering
\subfloat[ROC curves on TJU\_PV dataset]{
		\includegraphics[width=0.3\linewidth]{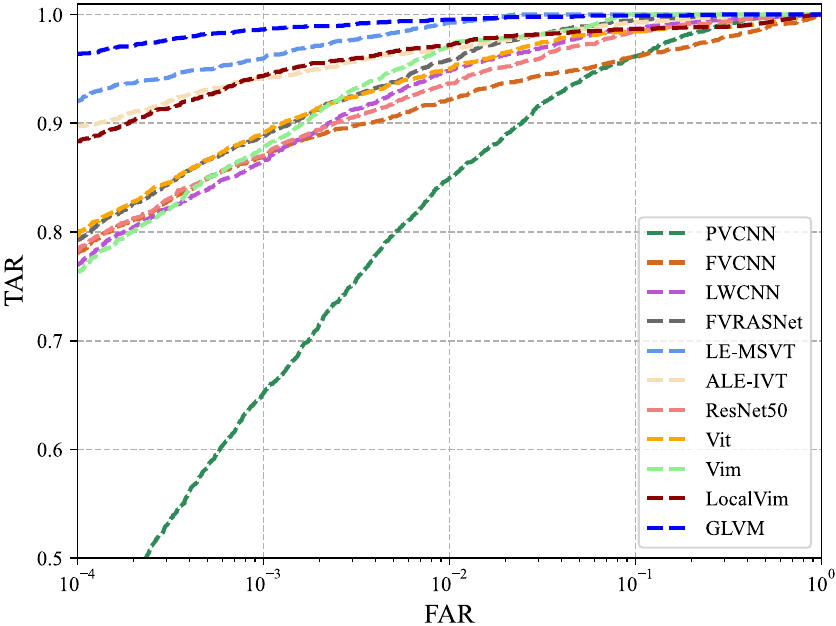}}
\subfloat[ROC curves on HKPU\_PV dataset dataset]{
		\includegraphics[width=0.3\linewidth]{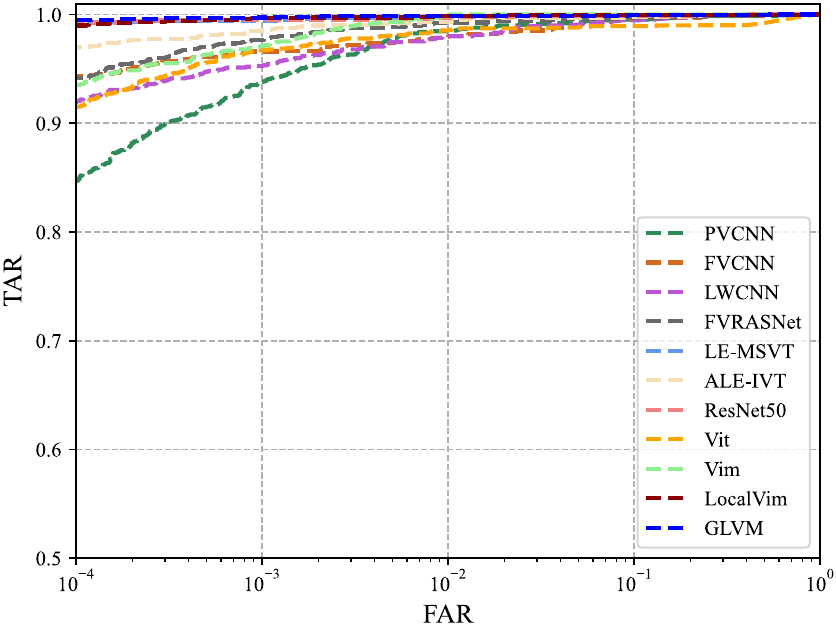}}
\subfloat[ROC curves on VERA\_PV dataset]{
		\includegraphics[width=0.3\linewidth]{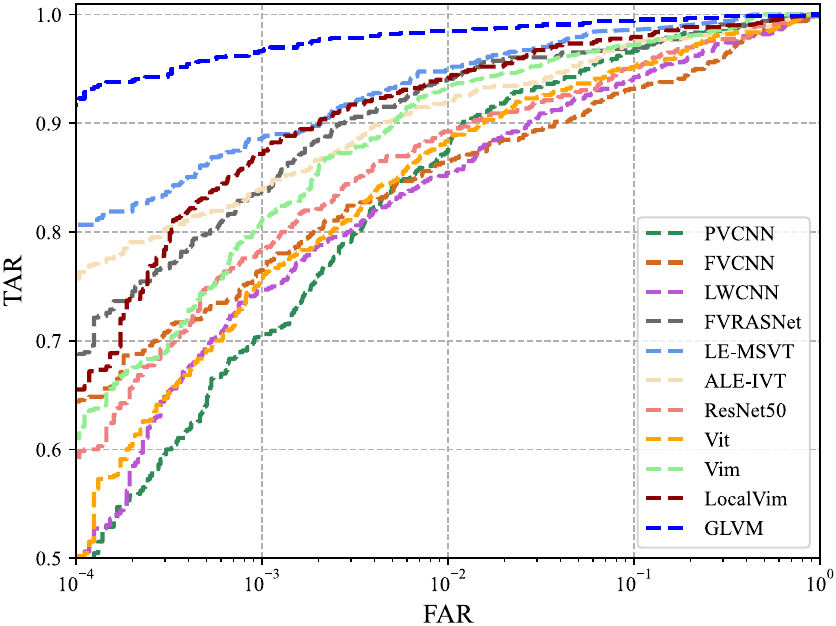}}
\caption{ROC curves for different models on three public vein datasets.}
\label{roc_fit1}
\end{figure*}

To benchmark our proposed Global-local Vision Mamba (GLVM), we implement state of art models, namely (\textit{i.e.} PVCNN \cite{9354642}, FVCNN \cite{das2018convolutional}, LWCNN \cite{shen2021finger}, FVRASNet \cite{yang2020fvras}, LE-MSVT \cite{qin2023label}, ALE-IVT \cite{qin2024attention}, ResNet50 \cite{He_2016_CVPR}, Vim 
 \cite{zhu2024vision}, Vit \cite{ dosovitskiy2020image}, LocalVim \cite{huang2024localMamba}), trained end-to-end from scratch. The architecture hyperparameters in Table \ref{ss} for our GLVM are determined by prior knowledge. Specifically, we set  $N=12$, $C=64$, $D=32$, $T=4$, $E=2$, and $V=16$. The recognition accuracy (ACC) and equal error rate (EER) for the above approaches on the three datasets are illustrated in Table \ref{acc}, with the corresponding ROC curves shown in Fig. \ref{roc_fit1}.
\begin{table}[h!]
    \centering
    \caption{Recognition accuracy (\%) and EER (\%) of various approaches on three public datasets.}
    \setlength{\abovecaptionskip}{1pt}
    \setlength{\belowcaptionskip}{1pt}
    \belowrulesep=0pt
    \aboverulesep=0pt
    \renewcommand\arraystretch{1.3}
    \scalebox{0.93}{
        \begin{tabular}{c|cc|cc|cc}
        \toprule[1.1pt]
        \multirow{2}{*}{Approach}           & \multicolumn{2}{c|}{TJU\_PV}        & \multicolumn{2}{c|}{HKPU\_PV}        & \multicolumn{2}{c}{VERA\_PV} \\
        \cline{2-7}
                                            & ACC  & EER    & ACC  & EER     & ACC  & EER           \\
        \midrule[0.9pt]
        PVCNN\cite{9354642}                 & 80.52       & 3.57          & 95.80       & 1.66            & 85.30       & 3.19          \\
        FVCNN\cite{das2018convolutional}    & 86.43       & 3.03          & 94.87       & 1.82            & 79.58       & 4.58          \\
        LWCNN\cite{shen2021finger}          & 85.64       & 3.09          & 93.47       & 1.73            & 80.76       & 4.74          \\
        FVRASNet\cite{yang2020fvras}        & 87.33       & 2.29          & 94.67       & 0.85            & 89.12       & 2.81          \\
        LE-MSVT\cite{qin2024attention}      & 92.57  & 1.23     &96.87     & 0.72     & 84.55    & 3.63           \\
        ALE-IVT\cite{qin2023label}   & \textcolor{cyan}{93.63}   & \textcolor{cyan}{0.87}    & \textcolor{cyan}{99.13}    & \textcolor{cyan}{0.26}   & \textcolor{cyan}{91.52}   & \textcolor{cyan}{2.09}          \\
        ResNet50\cite{He_2016_CVPR}         & 90.23       & 1.47          & 98.93       & 0.39            & 89.55       & 2.48          \\
        Vit\cite{ dosovitskiy2020image}     & 84.40    & 3.26       & 94.00   & 1.05      & 77.12    & 4.83        \\
        Vim\cite{zhu2024vision}             & 85.57       & 2.86          & 95.40       & 0.99            & 82.79       & 3.92          \\
        LocalVim\cite{huang2024localMamba}  & 91.47       & 1.34          & 98.47       & 0.32             & 86.97       &  2.76         \\
        \textbf{GLVM}                  & \textbf{95.57}       & \textbf{0.47}          & \textbf{99.40}       & \textbf{0.20}            & \textbf{94.23}       & \textbf{0.62}          \\
        \midrule[0.9pt]
        Gain                                & \textbf{\textcolor{ForestGreen}{+1.94}}  & \textbf{\textcolor{ForestGreen}{-0.40}} & \textbf{\textcolor{ForestGreen}{+0.27}} & \textbf{\textcolor{ForestGreen}{-0.06}} & \textbf{\textcolor{ForestGreen}{+2.71}} & \textbf{\textcolor{ForestGreen}{-1.47}} \\
        \bottomrule[1.2pt]
        \end{tabular}
    }
    \label{acc}
\end{table}

From Table \ref{acc}, our approach outperforms the state of the art approaches above by obtaining the highest recognition accuracy and lowest verification errors.  Specifically,  our GLVM's accuracy is $95.57\%$, $99.40\%$ and $94.23\%$ on TJU\_PV, HKPU\_PV, and VERA\_PV datasets, respectively, which are $1.94\%$, $0.27\%$ and $2.71\%$ higher than the best benchmarking method. Also, our method achieves the lowest EER, $0.47\%$, $0.20\%$, and $0.62\%$ on the three datasets and outperforms the benchmarking methods with an error reduction of $0.40\%$, $0.06\%$, and $1.47\%$. From Fig. \ref{roc_fit1}, we also observe that our method achieves the highest accuracy at different FARs, which may be attributed to the following facts: 
1) The Structured State Space Models (SSMs) in Mamba, enjoying linear complexity and a global receptive field, have shown superior performance compared to Transformers in some visual classification tasks \cite{zhu2024vision, huang2024localMamba}. Besides, we have investigated a Multi-head Mamba to further improve Mamba's feature representation capacity, as shown in  Fig. \ref{MHMambaBlock}. 2) Our ConvMamba block (Fig. \ref{ConvMambaBlock}) in GLVM consists of Mamba and CNN modules. The convolution operators extract local features while  Manmba captures global representations \cite{zhu2024vision}. Our approach, therefore, exploits the complementarity of local features and global representations to ensure a robust vein feature representation. For single branch-based approaches \textit{i.e.} single CNN branch (PVCNN \cite{9354642}, FVCNN \cite{das2018convolutional}, LWCNN \cite{shen2021finger}, FVRASNet \cite{yang2020fvras}, ResNet50 \cite{He_2016_CVPR}), single Transformer branch (LE-MSVT \cite{qin2023label}, Vit \cite{ dosovitskiy2020image}) and single Mamba branch (Vim \cite{zhu2024vision}, LocalVim \cite{huang2024localMamba}) show poor performance in our experiments. This may be attributed to the following facts: 1) Although CNNs effectively capture local features by sharing weights and local perception, they fail to capture long-distance relationships between visual elements due to their limited receptive field \cite{li2023local,guo2022cmt}; 2) Transformer and Mamba, by contrast, enlarge the receptive fields, allowing for global representations with long-distance feature dependencies \cite{li2023local,zhu2024vision,huang2024localMamba,guo2022cmt} but they ignore local feature details. ALE-IVT \cite{qin2024attention} combines the CNN and Transformer to learn local features and global representations, thereby achieving higher recognition performance.

\begin{figure*}[ht!]
\centering
\subfloat[ROC curves on TJU\_PV dataset]{
		\includegraphics[width=0.3\linewidth]{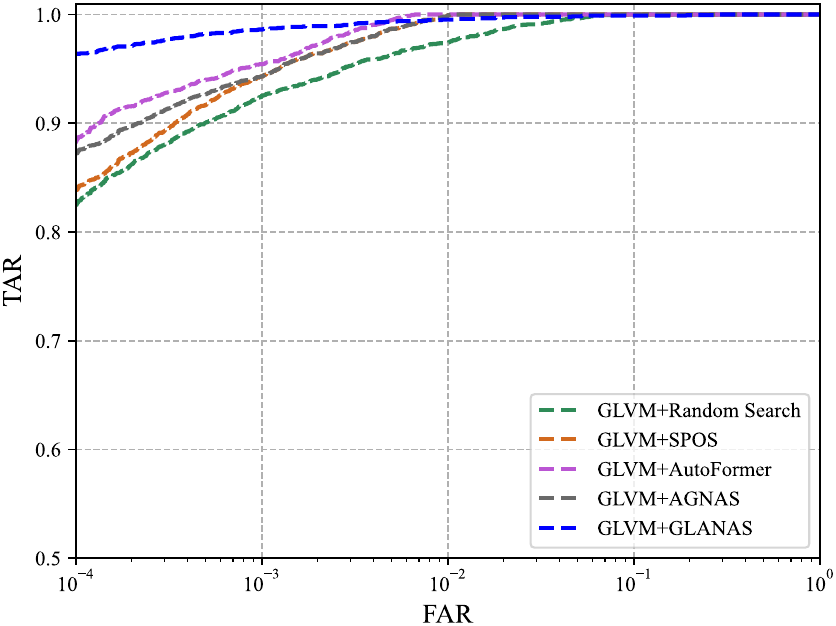}}
\subfloat[ROC curves on HKPU\_PV dataset dataset]{
		\includegraphics[width=0.3\linewidth]{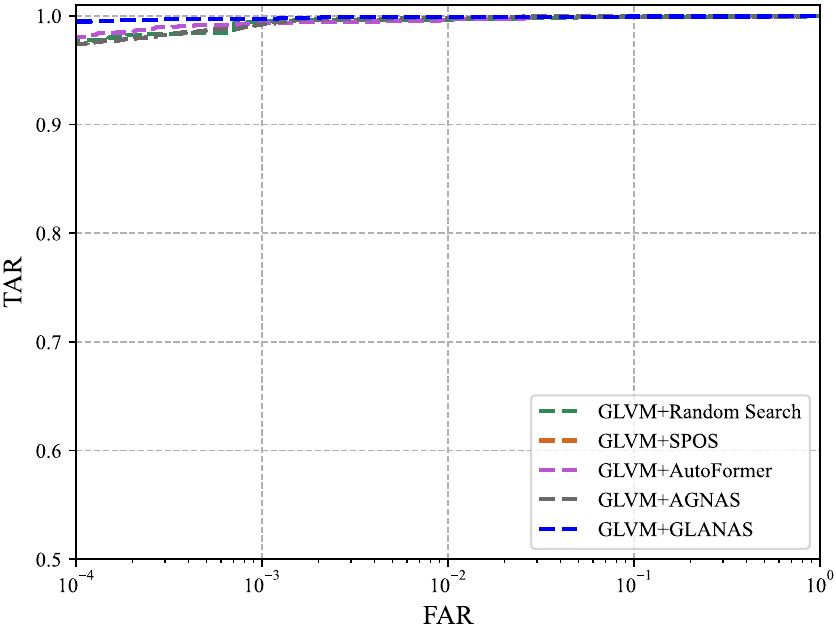}}
\subfloat[ROC curves on VERA\_PV dataset]{
		\includegraphics[width=0.3\linewidth]{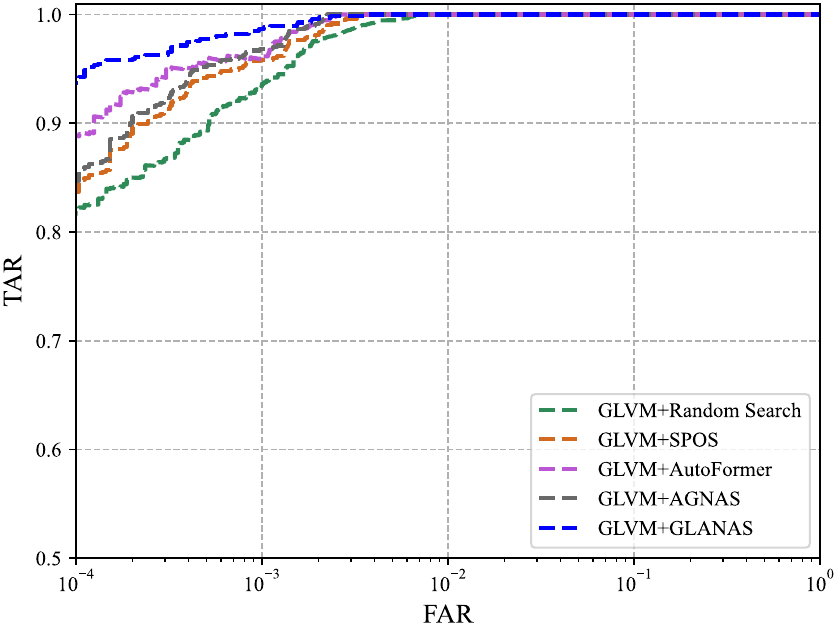}}
\caption{ ROC curves for different NAS methods on three public vein datasets.}
\label{roc_fit2}
\end{figure*}

\subsection{Performance of GLANAS} \label{poa}
To further assess performance, we compare our GLANAS approach with NAS approaches, namely Random Search \cite{bergstra2012random}, SPOS \cite{guo2020single}, AutoFormer \cite{chen2021autoformer} and AGNAS \cite{qin2023ag}. For fair comparison, all methods are used to search GLVM's hyperparameters with the same search space (Table \ref{ss}). The results are listed in Table \ref{nas_acc} while the ROC curves in Fig. \ref{roc_fit2}.
% All search methods are used in our proposed search space. For fair comparisons, we train all the supernets of comparison methods for 500 epochs, which is the same as the training epochs in ours. For our GLANAS method, the number of alternating searches is 3. After supernet training, we select 5 architectures and retrain them for SPOS, Weight Entanglement, and our methods. For AGNAS, the controller generates an optimal network to retrain for final evaluation. For the random search, we randomly sample and retrain 5 networks. The architecture with the highest retraining accuracy in the validation set is chosen as the final model. 
\begin{table}[h!]
    \centering
    \caption{Performance comparisons between different search methods on three public datasets.}
    \setlength{\abovecaptionskip}{1pt}
    \setlength{\belowcaptionskip}{1pt}
    \belowrulesep=0pt
    \aboverulesep=0pt
    \renewcommand\arraystretch{1.3}
    \scalebox{0.85}{
        \begin{tabular}{c|cc|cc|cc}
        \toprule[1.1pt]
        \multirow{2}{*}{Approach}           & \multicolumn{2}{c|}{TJU\_PV}        & \multicolumn{2}{c|}{HKPU\_PV}        & \multicolumn{2}{c}{VERA\_PV} \\
        \cline{2-7}
                                            & ACC  & EER    & ACC  & EER     & ACC  & EER           \\
        \midrule[0.9pt]
        GLVM+Random Search\cite{bergstra2012random}               & 94.90   & 0.55   & 99.34   & 0.27   & 94.13   & 0.75         \\
       GLVM+ SPOS\cite{guo2020single}                             & 95.63   & 0.48   & 99.47   & 0.31   & 94.53   & 0.68         \\
        GLVM+AutoFormer\cite{chen2021autoformer}         & \textcolor{cyan}{95.78}   & 0.44   & \textcolor{cyan}{99.54}   & 0.22   & \textcolor{cyan}{94.79}   & \textcolor{cyan}{0.59}         \\
        GLVM+AGNAS\cite{qin2023ag}                                & 95.74   & \textcolor{cyan}{0.42}   & 99.45   & \textcolor{cyan}{0.16}   & 94.68   & 0.65     \\
        \textbf{GLVM+GLANAS}               & \textbf{96.84}       & \textbf{0.27}          & \textbf{99.63}       & \textbf{0.07}            & \textbf{95.87}       & \textbf{0.48}          \\
        \midrule[0.9pt]
        Gain                                & \textbf{\textcolor{ForestGreen}{+1.06}}  & \textbf{\textcolor{ForestGreen}{-0.15}} & \textbf{\textcolor{ForestGreen}{+0.09}} & \textbf{\textcolor{ForestGreen}{-0.09}} & \textbf{\textcolor{ForestGreen}{+1.08}} & \textbf{\textcolor{ForestGreen}{-0.11}} \\
        \bottomrule[1.2pt]
        \end{tabular}
    }
    \label{nas_acc}
\end{table}

From Table \ref{nas_acc} and Fig. \ref{roc_fit2}, our GLVM with GLANAS significantly improves accuracy w.r.t other approaches, namely + $1.06\%$, $0.09\%$, and $1.08\%$  \textit{w.r.t} the second best approach on the TJU\_PV, HKPU\_PV, and VERA\_PV datasets, respectively. Applying GLANAS to search the GLVM architecture reduces EER by $0.15\%$, $0.09\%$, and $0.11\%$ on the three datasets, respectively. Random Search \cite{bergstra2012random} achieves the worst performance, as it fails, due to its uncertainty, to select a good architecture in huge search spaces.  
{SPOS \cite{guo2020single} constructs a simplified supernet, where all subnets share the weights in their modules. As such an architecture search is extremely memory-consuming and slow for large datasets, it achieves poor performance.} 
Similarly, the reinforcement learning proposed to sample a subnet from the supernet \cite{qin2023ag} for vein recognition shows slow convergence and low performance, as the weights of its different blocks in the same layer are independent. The weight entanglement strategy of AutoFormer \cite{chen2021autoformer}, by contrast, enables different blocks to share weights for their common parts in each layer, achieving thereby higher performance than weight sharing based approaches \cite{qin2023ag}  \cite{guo2020single}. As described in Section \ref{ssc}, however, the GLVM search space is huge, implying low correlation between the retrained subnet and the subnet sampled from the supernet. These supernet-based architecture search approaches \cite{guo2020single,qin2023ag,chen2021autoformer}, therefore, are unreliable. Our GLANAS divides the huge space into an effective search space range of existing NAS methods, making it easier for weight entanglement-based NAS \cite{chen2021autoformer} to obtain better models.

In addition, comparing the results in Table \ref{acc} and Table \ref{nas_acc}, we found that the resulting GLVM achieves better performance after automatically searching its architecture instead of manually selecting it.  The reason is that existing NAS and our GLANAS optimize the architecture of our GLVM for different tasks, eliminating the risk of manually discarding optimal networks for classification. Similar conclusions have been demonstrated in recent works \cite{qin2023ag,gong2022nasvit}.

\begin{table*}[h!]
    \centering
    \caption{Searching architecture on three public datasets.}
    \setlength{\abovecaptionskip}{1pt}
    \setlength{\belowcaptionskip}{1pt}
    \belowrulesep=0pt
    \aboverulesep=0pt
    \renewcommand\arraystretch{1.3}
    \scalebox{0.8}{
        \begin{tabular}{|c|c|c|c|c|c|c|c|c|c|c|c|c|c|}
        \toprule[0.6pt]
        \multirow{2}{*}{dataset}       & \multirow{2}{*}{Global Architecture} & \multicolumn{12}{c|}{Local Architecture} \\ 
        \cline{3-14}
                                        &                                      & 1&2&3&4&5&6&7&8&9&10&11&12        \\ 
        \midrule[0.6pt]
        \multirow{1}{*}{TJU\_PV}        & (9,320,128)  &(4,2,16)&(6,1.5,32)&(4,2,16)&(4,2.5,32)&(2,1.5,48)&(4,1,48)&(6,1,48)&(2,2,16)&(4,2.5,32)&&&        \\ 
        \midrule[0.6pt]
        \multirow{1}{*}{HKPU\_PV}       & (9,256,64)   &(2,1,48)&(2,2,32)&(2,2,48)&(4,1.5,16)&(4,1.5,48)&(2,2.5,48)&(4,1.5,16)&(2,1,32)&(4,2,16)&&&        \\ 
        \midrule[0.6pt]
        \multirow{1}{*}{VERA\_PV}       & (6,192,64)   &(4,2,32)&(4,2,16)&(4,1,16)&(2,2,16)&(4,1,48)&(4,1,16)&&&&&&        \\ 
        \bottomrule[0.6pt]
        \end{tabular}
    }
    \label{sr}
\end{table*}

\begin{figure*}[ht]
    \centering
    \includegraphics[width=0.8\linewidth]{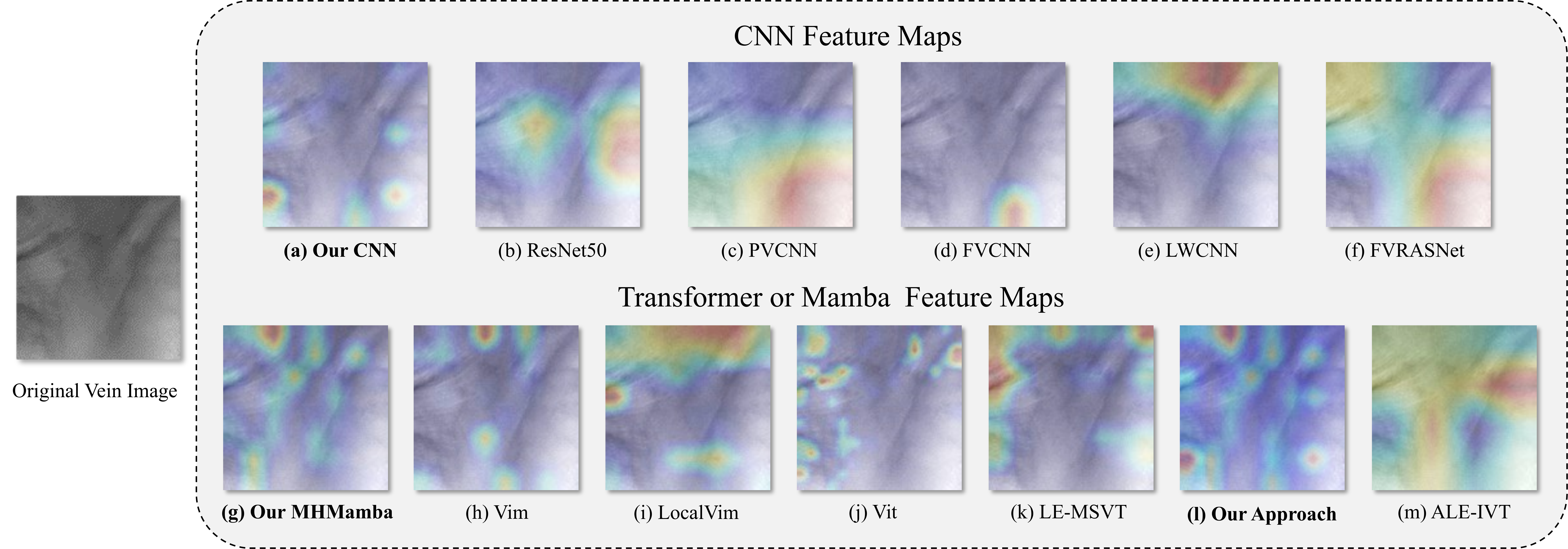}
    \caption{Feature Heat maps in various methods by using the GradCAM method \cite{selvaraju2020grad}. The heat maps from (a) CNN branch in our approach, (b) ReNet50 \cite{He_2016_CVPR}, (c) PVCNN \cite{9354642}, (d) FVCNN \cite{das2018convolutional}, (e) LWCNN \cite{shen2021finger}, (f) FVRASNet \cite{yang2020fvras}, (g) MHMamba branch in our approach, (h) Vim \cite{zhu2024vision}, (i) LocalVim \cite{huang2024localMamba}, (j) Vit \cite{dosovitskiy2020image}, (k) LE-MSVT \cite{qin2023label}, (l) Our Approach, and (m) ALE-IVT \cite{qin2024attention}. (Best viewed in color)}
    \label{visual}
\end{figure*}

\subsection{Ablation Experiments}
\textbf{(1) GLVM Architecture Ablation.}
~The ConMamba block in GLVM (Fig. \ref{ConvMambaBlock}) consists of three branches: Multi-head Mamba, CNN, and Feature Interaction Unit. This experiment investigates the impact of different components on performance. 

In Table \ref{sr},  first, only the Multi-head Mamba branch is retrained in the obtained architectures, which is represented by the \textit{Multi-head Mamba branch} in Table \ref{modelablation}. Second, we only retain the CNN branch and denote it as the \textit{CNN branch}. Third, we replace the Multi-head Mamba block in the Multi-head Mamba branch with a single-head Mamba block, (\textit{i.e.} original Mamba block \cite{zhu2024vision}), with the resulting model denoted as \textit{Single-head Mamba branch}. Fourth, we combine prediction scores of the CNN branch and Multi-head Mamba branch (without Feature Interaction Unit), with such a scheme named \textit{Dual branch}. Finally, the Feature Interaction Unit branch is used to fuse the CNN branch and Multi-head Mamba branch, with the resulting model denoted \textit{Dual branch + FIU }, i.e. GLVM. The results in Table. \ref{modelablation} show that the integration of the Multi-head Mamba branch, CNN branch, and Feature Interaction Unit branch allows outperforms all methods, including single-head Mamba, and achieves the highest recognition accuracy.

\begin{table}[H]
    \centering
    \caption{Ablation experiment results of proposed GLVM on three public datasets.}
    \setlength{\abovecaptionskip}{1pt}
    \setlength{\belowcaptionskip}{1pt}
    \belowrulesep=0pt
    \aboverulesep=0pt
    \renewcommand\arraystretch{1.3}
    \scalebox{0.85}{
        \begin{tabular}{c|cc|cc|cc}
        \toprule[1.0pt]
        \multirow{2}{*}{Approach}   & \multicolumn{2}{c|}{TJU\_PV}   & \multicolumn{2}{c|}{HKPU\_PV}  & \multicolumn{2}{c}{VERA\_PV} \\
        \cline{2-7}
                                    & ACC  & EER   & ACC  & EER   & ACC  & EER           \\
        \midrule[0.8pt]
        Single-head Mamba branch          & 85.57    & 2.86   & 95.40   & 0.99   & 82.79   & 3.92          \\
        Multi-head Mamba branch          & 89.13    & 1.63   & 96.33   & 0.64   & 85.97   & 2.61          \\
        CNN branch                & 93.36    & 0.96   & 97.73   & 0.54   & 92.58   & 1.89          \\
        Dual branch                 & 94.54    & 0.89   & 98.33   & 0.28   & 93.09   & 1.22          \\      
        \textbf{Dual branch + FIU (GLVM)}      & \textbf{95.57}  & \textbf{0.47}  & \textbf{99.40} & \textbf{0.20}  & \textbf{94.23}  & \textbf{0.62}   \\
        \bottomrule[1.0pt]
        \end{tabular}
    }
    \label{modelablation}
\end{table}

\begin{table}[H]
    \centering
    \caption{Ablation experiment results of proposed GLANAS on three public datasets.}
    \setlength{\abovecaptionskip}{1pt}
    \setlength{\belowcaptionskip}{1pt}
    \belowrulesep=0pt
    \aboverulesep=0pt
    \renewcommand\arraystretch{1.3}
    \scalebox{0.80}{
        \begin{tabular}{c|cc|cc|cc}
        \toprule[1.0pt]
        \multirow{2}{*}{Approach}   & \multicolumn{2}{c|}{TJU\_PV}   & \multicolumn{2}{c|}{HKPU\_PV}  & \multicolumn{2}{c}{VERA\_PV} \\
        \cline{2-7}
                                    & ACC  & EER    & ACC  & EER   & ACC  & EER           \\
        \midrule[0.8pt]                                 
       % Random Search        & 92.90    &  1.17   &  98.04   & 0.27   & 90.42   & 2.65         \\
        %One Stage with Weight Sharing            & 93.63    &  1.05   &  98.27   & 0.41   & 91.98   & 2.58          \\
        One stage searching          & 95.78    &  0.44   &  99.54   & 0.22   & 94.79   & 0.59         \\
        Two stage searching                 & 96.27    &  0.36   &  99.58   & 0.14   & 94.83   & 0.56         \\
        \textbf{Alternative searching (GLANAS)}    & \textbf{96.84}   & \textbf{0.27}     & \textbf{99.63}  & \textbf{0.07}   & \textbf{95.30}  & \textbf{0.48}   \\

        \bottomrule[1.0pt]
        \end{tabular}
    }
    \label{NASablation}
\end{table}
\textbf{(2) GLANAS Ablation.} 
~Our GLANAS alternatively optimizes the architecture on the local search stage and global search stage, with the weight entanglement strategy   \cite{chen2021autoformer} used to find the optimal parameters at each stage. 
This strategy  \cite{chen2021autoformer} is first directly used to find the optimal hyperparameters on the whole search space (Table \ref{ss}), which is denoted as \textit{One stage searching}. The search space is then divided into global and local spaces by our GLANAS and we gradually perform global and local search for one time. This scheme is represented by the \textit{Two stage searching}. Finally, we alternatively search the optimal hyperparameters on the two spaces until convergence, which is denoted as \textit{Alternative searching (GLANAS)}. The ablation results in Table \ref{NASablation}
show that the alternative searching on the two stages significantly improves performance, which demonstrates that the space division and the alternating searching scheme are helpful for NAS approaches to find the optimal architecture hyperparameters.

\subsection{Feature Visualization Analysis}
To qualitatively assess our approach performance, we visualize the heat maps of the last hidden layers of the various models. As our GLVM includes the CNN and MHMamba branches, we show the CNN, MHMamba, and dual-branch hybrid feature maps. Similarly, the CNN-based feature maps in ResNet50 \cite{He_2016_CVPR}, PVCNN \cite{9354642}, FVCNN \cite{das2018convolutional}, LWCNN \cite{shen2021finger}, FVRASNet \cite{yang2020fvras}, Transformer-based feature maps in Vit \cite{dosovitskiy2020image}, LE-MSVT \cite{qin2023label}, ALE-IVT \cite{qin2024attention}, and Mamba-based feature maps in Vim \cite{zhu2024vision}, LocalVim \cite{huang2024localMamba}, are extracted. To facilitate comparison, we normalize the output maps in each channel to the same size as the input image and average the resulting maps of all channels to obtain the final heat map. The results are shown in Fig. \ref{visual}. 
From the feature maps in Figs. \ref{visual}(a)-(f), we observe that CNN models focus on a few regions with high brightness in a palm image. The single Mamba and Transformer models, by contrast, focus on texture feature regions, distributed at different locations in Figs. \ref{visual}(g)-(k). We observe also that our MHMamba branch (Fig. \ref{visual}(g)) focuses more, w.r.t other approaches, on object regions such as vein textures, which hints to the richest feature representation inferred by our MHMamba. Finally, compared to two local and global approaches, \textit{i.e.} ALE-IVT (Fig. \ref{visual}(m)) and our GLVM (Fig. \ref{visual}(l)), we clearly observe that our approach focuses more precisely on vein textures.

\section{Conclusion}
For vein recognition, we have proposed in this paper a Global-Local Vision Mamba to take advantage of both local receptive field-based convolutions and state space models for enhanced representation learning. Second, we have propose a Global-Local Alternating Neural Architecture Search method to find optimal architecture hyperparameters. The experimental results on three public palm-vein datasets show that our approach significantly outperforms state-of-the-art approaches. In the future, we intend to explore the application of Mamba to other biometrics such as finger-vein recognition and face recognition.

\bibliographystyle{unsrt}
\bibliography{bibtex}
\end{document}